\documentclass[10pt]{article} 
\usepackage[preprint]{tmlr}

\usepackage{amsmath,amsfonts,bm}

\def\eqref#1{equation~\ref{#1}}

\def\1{\bm{1}}

\DeclareMathAlphabet{\mathsfit}{\encodingdefault}{\sfdefault}{m}{sl}
\SetMathAlphabet{\mathsfit}{bold}{\encodingdefault}{\sfdefault}{bx}{n}

\usepackage{hyperref}
\usepackage{url}

\usepackage{graphicx}
\usepackage{overpic}
\usepackage{booktabs}
\usepackage[detect-all]{siunitx}
\newcommand{\code}[1]{\texttt{#1}}

\title{Analyzing Deep Transformer Models for Time Series Forecasting via Manifold Learning}

\author{\name Ilya Kaufman \email ilyakau@post.bgu.ac.il \\
      \addr Ben-Gurion University of the Negev\\
      \name Omri Azencot \email azencot@cs.bgu.ac.il \\
      \addr Ben-Gurion University  of the Negev\\}

\begin{document}

\maketitle

\begin{abstract}

Transformer models have consistently achieved remarkable results in various domains such as natural language processing and computer vision. However, despite ongoing research efforts to better understand these models, the field still lacks a comprehensive understanding. This is particularly true for deep time series forecasting methods, where analysis and understanding work is relatively limited. Time series data, unlike image and text information, can be more challenging to interpret and analyze. To address this, we approach the problem from a \emph{manifold learning} perspective, assuming that the latent representations of time series forecasting models lie next to a low-dimensional manifold. In our study, we focus on analyzing the geometric features of these latent data manifolds, including intrinsic dimension and principal curvatures. Our findings reveal that deep transformer models exhibit similar geometric behavior across layers, and these geometric features are correlated with model performance. Additionally, we observe that untrained models initially have different structures, but they rapidly converge during training.
By leveraging our geometric analysis and differentiable tools, we can potentially design new and improved deep forecasting neural networks. This approach complements existing analysis studies and contributes to a better understanding of transformer models in the context of time series forecasting. Code is released at \url{https://github.com/azencot-group/GATLM}. 

\end{abstract}

\section{Introduction}

Over the past decade, modern deep learning has shown remarkable results on multiple challenging tasks in computer vision~\citep{krizhevsky2012imagenet}, natural language processing (NLP)~\citep{pennington2014glove}, and speech recognition~\citep{graves2013speech}, among other domains~\citep{berman2023multifactor, naiman2024generative, goodfellow2016deep}. Recently, the transformer~\citep{vaswani2017attention} has revolutionized NLP by allowing neural networks to capture long-range dependencies and contextual information effectively. In addition, transformer-based architectures have been extended to non-NLP fields, and they are among the state-of-the-art (SOTA) models for vision~\citep{dosovitskiy2020image} as well as time series forecasting (TSF)~\citep{wu2021autoformer, zhou2022fedformer}. Unfortunately, while previous works, e.g., \citet{zeiler2014visualizing, karpathy2015visualizing, tsai2019transformer} among others, attempted to explain the underlying mechanisms of neural networks, deep transformer models are still considered not well understood.

The majority of approaches analyzing the inner workings of vision and NLP transformer models investigate their attention modules~\citep{bahdanau2014neural} and salient inputs~\citep{wallace2019allennlp}. Unfortunately, time series forecasting methods have received significantly less attention. This may be in part due to their relatively recent appearance as strong contenders on TSF in comparison to non-deep and hybrid techniques~\citep{oreshkin2019n}. Further, while vision and NLP modalities may be ``natural'' to interpret and analyze~\citep{naiman2023operator}, time series requires analysis tools which may be challenging to develop for deep models. For instance, N-BEATS~\citep{oreshkin2019n} designed a method that promotes the learning of trend and seasonality parts, however, their model often recovers latent variables whose relation to trend and seasonality is unclear~\citep{challu2022n}. Moreover, there is already a significant body of work of SOTA TSF that warrants analysis and understanding. Toward bridging this gap, we investigate in this work the geometric properties of latent representations of transformer-based TSF techniques via \emph{manifold learning} tools.

Manifold learning is the study of complex data encodings under the \emph{manifold hypothesis} where high-dimensional data is assumed to lie close to a low-dimensional manifold~\citep{coifman2006diffusion}. This assumption is underlying the development of numerous machine learning techniques, akin to considering independent and identically distributed (i.i.d.) samples~\citep{goodfellow2016deep}. Recent examples include works on vision~\citep{nguyen2019neural}, NLP~\citep{hashimoto2016word}, and time series forecasting~\citep{papaioannou2022time}. However, to the best of our knowledge, there is no systematic work that analyzes transformer-based TSF deep neural networks from a manifold learning perspective. In what follows, we advocate the study of geometric features of Riemannian manifolds~\citep{lee2006riemannian} including their \emph{intrinsic dimension} (ID) and \emph{mean absolute principal curvature} (MAPC). The ID is the minimal degrees of freedom needed for a lossless encoding of the data, and MAPC measures the deviation of a manifold from being flat.

Previously, geometric features of data manifolds were considered in the context of analyzing deep convolutional neural networks (CNN)~\citep{ansuini2019intrinsic, kaufman2023data}. Motivated by these recent works, we extend their analysis on image classification to the time series forecasting setting, focusing on SOTA TSF models~\citep{wu2021autoformer, zhou2022fedformer} evaluated on several multivariate time series datasets. We aim at characterizing the dimension and curvature profiles of latent representations along layers of deep transformer models. Our study addresses the following questions: (i) how do dimensionality and curvature change across layers? are the resulting profiles similar for different architectures and datasets? (ii) is there a correlation between geometric features of the data manifold to the performance of the model? (iii) how do untrained manifolds differ from trained ones? how do manifolds evolve during training?

Our results show that transformer forecasting manifolds undergo two phases: during encoding, curvature and dimensionality either drop or stay fixed, and then, during the decoding part, both dimensionality and curvature increase significantly. Further, this behavior is shared across several architectures, datasets and forecast horizons. In addition, we find that the MAPC estimate is correlated with the test mean squared error, allowing one to compare models without access to the test set. Moreover, this correlation is unlike the one found in deep neural networks for classification. Finally, untrained models show somewhat random dimension and curvature patterns, and moreover, geometric manifolds converge rapidly (within a few epochs) to their final geometric profiles. This finding may be related to studies on the neural tangent kernel~\citep{li2018learning, jacot2018neural} and linear models for forecasting~\citep{zeng2023transformers}. We believe that our geometric insights, results and tools may be used to design new and improved deep forecasting tools.

\section{Related Work}

Our research lies at the intersection of understanding deep transformer-based models, and manifold learning for analysis and time series. We focus our discussion on these topics.
 
\paragraph{Analysis of transformers.} Large transformer models have impacted the field of NLP and have led to works such as \citet{vig2019multiscale} that analyze the multi-head attention patterns and found that specific attention heads can be associated with various grammatical functions, such as co-reference and noun modifiers. Several works~\citep{clark2019does, tenney2019bert, rogers2021primer} study the BERT model~\citep{devlin2019bert} and show that lower layers handle lexical and syntactic information such as part of speech, while the upper layers handle increasingly complex information such as semantic roles and co-reference. In~\citet{dosovitskiy2020image}, the authors inspect patch-based vision transformers (ViT) and find that the models globally attend to image regions that are semantically relevant for classification. \citet{caron2021emerging} show that a self-supervised trained ViT produces explicit representations of the semantic location of objects within natural images. \citet{chefer2021transformer} compute a relevancy score for self-attention layers that is propagated throughout the network, yielding a visualization that highlights class-specific salient image regions. \citet{nie2023time} recently studied the effectiveness of transformer in TSF in terms of their ability to extract temporal relations, role of self-attention, temporal order preservation, embedding strategies, and their dependency on train set size. While in general they question the effectivity of transformer for forecasting, new transformer-based approaches continue to appear~\citep{zeng2023transformers}, consistently improving the state-of-the-art results on common forecasting benchmarks.

\paragraph{Manifold learning analysis.} Motivated by the ubiquitous manifold hypothesis, several existing approaches investigate geometric features of data representations across different layers. In~\citet{hauser2017principles}, the authors formalize a Riemannian geometry theory of deep neural networks (DNN) and show that residual neural networks are finite difference approximations of dynamical systems. \citet{yu2018curvature} compare two neural networks by inspecting the Riemann curvature of the learned representations in fully connected layers. \citet{cohen2020separability} examine the dimension, radius and capacity throughout the training process, and they suggest that manifolds become linearly separable towards the end of the layer's hierarchy. \citet{doimo2020hierarchical} analyzed DNNs trained on ImageNet and found that the probability density of neural representations across different layers exhibits a hierarchical clustering pattern that aligns with the semantic hierarchy of concepts.  \citet{stephenson2021geometry} conclude that data memorization primarily occurs in deeper layers, due to decreasing object manifolds' radius and dimension, and that generalization can be restored by reverting the weights of the final layers to an earlier epoch. Perhaps closest to our approach are the works by \citet{ansuini2019intrinsic} and \citet{kaufman2023data}, where the authors estimate the intrinsic dimension and Riemannian curvature, respectively, of popular deep convolutional neural networks. Both works showed characteristic profiles and a strong correlation between the estimated geometric measure and the generalization error. Recently, \citet{valeriani2023geometry} investigated the intrinsic dimension and probability density of large transformer models in the context of classification tasks on protein and genetic sequence datasets. Complementary to previous works, our study focuses on the setting of regression time series forecasting problems using multivariate real-world time series datasets.

\paragraph{Manifold learning in time series forecasting.}  Unfortunately, latent representations of deep TSF received less attention in the literature, and thus we discuss works that generally investigate TSF from a manifold learning perspective. \citet{papaioannou2022time} embed high-dimensional time series into a lower-dimensional space using nonlinear manifold learning techniques to improve forecasting. Similarly, \citet{han2018structured} proposed a novel framework, which performs nonuniform embedding, dynamical system revealing, and time-series prediction. In~\citet{li2021combination}, the authors exploit manifold learning to extract the low-dimensional intrinsic patterns of electricity loads, to be used as input to recurrent modules for predicting low-dimensional manifolds. \citet{lin2006learning} employ a dynamic Bayesian network to learn the underlying nonlinear manifold of time series data, whereas \citet{shnitzer2017manifold} harness diffusion maps to recover the states of dynamical systems. \citet{kaufman2024first} approximate the data manifold to sample synthetic examples for regression problems. Finally, manifold-based distance functions for time series were proposed in~\citet{rodrigues2018multivariate, oReilly2017univariate}.

\section{Background and Method}

\paragraph{Time series forecasting.} Given a dataset of multivariate time series sequences $\mathcal{D} := \{ x_{1:T+h}^j \}_{j=1}^N$ where $x_{1:T+h} = x_1, \dots, x_{T+h} \subset \mathbb{R}^D$, the goal in time series forecasting (TSF) is to accurately forecast the series $x_{T+1:T+h}$, based on the sequence $x_{1:T}$, where we omit $j$ for brevity. The values $T$ and $h$ are typically referred to as lookback and horizon, respectively. The forecast accuracy can be measured in several ways of which the mean squared error (MSE) is the most common. We denote by $\tilde{x}_{T+1:T+h} = f(x_{1:T})$ the output of a certain forecast model, e.g., a neural network, then $e_\text{MSE} := \frac{1}{h} \sum_{t=T+1}^{T+h} \| x_t - \tilde{x}_t \|_2^2$ is the forecast error. In our study, we consider $T=96$, and $h=96, 192, 336$ and $720$, and standard benchmark datasets including Electricity, Traffic, ETTm1, ETTm2, ETTh1, ETTh2, and weather~\citep{wu2021autoformer}. In App.~\ref{app:tsf_datasets}, we provide a detailed description of the datasets and their properties.

\begin{figure*}[t]
  \centering
  \includegraphics[width=1\linewidth]{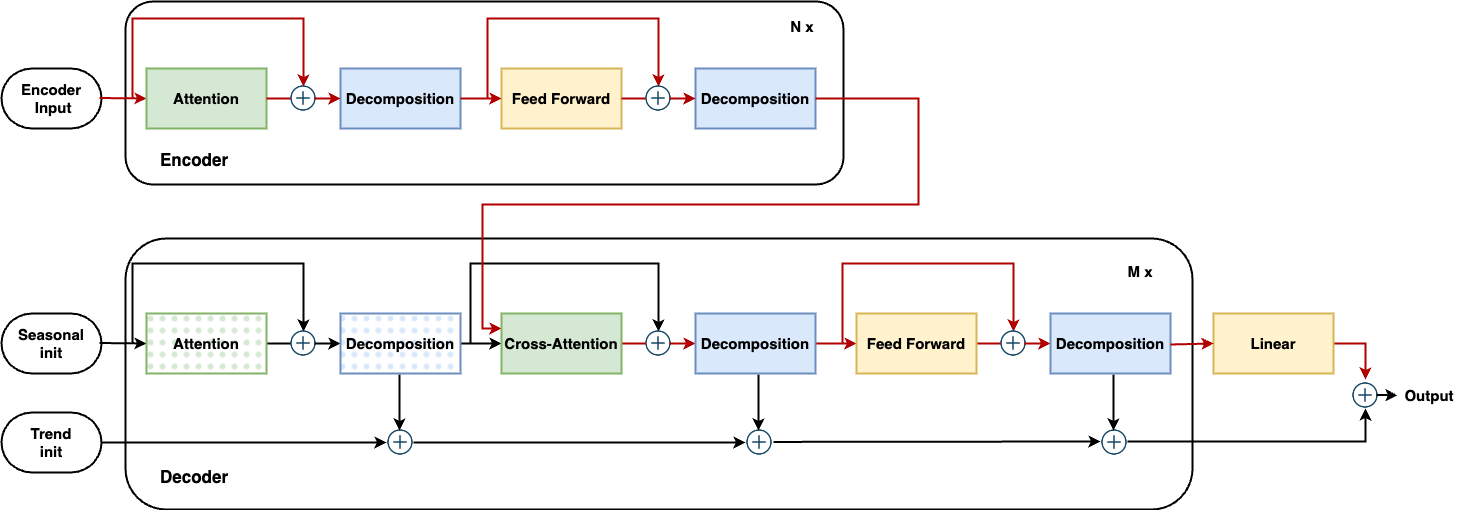}
  \caption{We study Transformer-based architectures~\citep{wu2021autoformer, zhou2022fedformer} that include two encoders and one decoder, and an output linear layer. We sample geometric features in the output of sequence decomposition layers, depicted as solid blue blocks.}
  \label{fig:arch}
\end{figure*}

\paragraph{Transformer-based TSF deep neural networks.} State-of-the-art (SOTA) deep time series forecasting models appeared only recently~\citep{oreshkin2019n}, enjoying a rapid development of transformer-based architectures, e.g.,~\citet{zhou2021informer, wu2021autoformer, liu2021pyraformer, zhou2022fedformer, nie2023time}, among many others. In what follows, we will focus on Autoformer~\citep{wu2021autoformer} and FEDformer~\citep{zhou2022fedformer} as they are established architectures that are still considered SOTA~\citep{nochumsohn2024data}. In App.~\ref{app:tsf_models}, we also mention additional TSF models and their analysis. Please see Fig.~\ref{fig:arch} for a schematic illustration of the architecture we investigate. The network is composed of two encoder blocks and a single decoder block, where the encoder and decoder blocks include two and three sequence decomposition layers, respectively. Both Autoformer and FEDformer utilize these decomposition layers to extract trend and seasonality information.
The network includes two inputs, the input signal $X$ which is fed to the encoder and the seasonality information $X_s$ of $X$ that serves as the input to the decoder. Note that the cross-attention module in the decoder receives two independent data streams, one is the output of the encoder while the other is a function of $X_s$.
Our goal is to analyze the geometrical features of the data propagation along the layers of the network in a serial manner (with respect to depth), therefore, we focus on the path from the encoder through the decoder to the output (shown in red in Fig.~\ref{fig:arch}). Namely, we follow data along the red trajectory, while ignoring the black trajectory. In particular, our analysis is based on sampling geometric properties of the data manifold after every decomposition module and after the final linear layer of the network. We chose the output of the decomposition blocks rather than the attention blocks since the Fourier Cross-correlation layer of the FEDformer model outputs almost identical values for all samples in the series, yielding zero curvature estimates.

\paragraph{Series decomposition.} While modern data analysis considers disentangled representations~\citep{naiman2023sample, berman2024sequential}, it is common in time series analysis to decompose signals to trend and seasonal parts~\citep{Anderson1976TimeSeries2E, cleveland1990stl}. This decomposition facilitates learning of complex temporal patterns, allowing to better capture global properties. In the time series forecasting domain, prior to Autoformer, time series decomposition was mainly used as a pre-processing tool, applied on the input data. Both Autoformer and FEDformer utilize decomposition as a core component, enabling it to progressively decompose the hidden time series throughout the entire forecasting process. This includes both historical (input) data and the predicted (output) intermediate results. The series decomposition block (noted as Decomposition in Fig.~\ref{fig:arch}) receives a time series signal as an input and outputs the seasonality (right arrow) and the trend (bottom arrow) such that their summation results in the original signal.

\paragraph{Encoder and decoder.} The encoder is utilized for modeling the seasonal parts as it discards the trend output from the decomposition block. The encoder's output, which includes past seasonal information, will serve as cross-information to assist the decoder in refining the prediction results. The decoder serves two purposes, it accumulates the trend components and propagates the seasonal component. The final prediction is the sum of the seasonal and trend components.

\paragraph{Geometric properties of data manifolds.} The fundamental assumption in our work is that data representations computed across layers of transformer-based models lie on Riemannian manifolds \citep{lee2006riemannian}. We are interested in computing the intrinsic dimension (ID) and the mean absolute principal curvature (MAPC) of the manifold, following recent work on deep CNNs~\citep{ansuini2019intrinsic, kaufman2023data}. We compute the ID using the TwoNN method~\citep{facco2017estimating} that utilizes the Pareto distribution of the ratio between the distances to the two closest neighbors to estimate the dimension. For the MAPC, we employ the curvature aware manifold learning (CAML) technique~\citep{li2018curvature} that parametrizes the manifold via its second-order Taylor expansion, allowing to estimate curvatures via the eigenvalues of local Hessian matrices. More details related to ID and MAPC are provided in App.~\ref{app:id_and_mapc}.

\paragraph{Data collection.} In this study, every architecture is trained on all datasets and horizons, using $10$ different seed numbers. For every combination of model, dataset, horizon and seed, we extract the latent data representations across layers, and we compute the ID and MAPC. The intrinsic dimension is estimated on $500$k point samples from $\mathcal{D}$, resulting in a single scalar value $d$. The estimated ID is used as an input to the CAML algorithm that uses $100$k samples, and it returns $d(D-d)$ principal curvatures per point, where $D$ is the extrinsic dimension. The MAPC of a manifold is calculated by computing the mean absolute value for each point and taking the mean over all points.

\begin{figure*}[t]
    \centering
    \includegraphics[width=1\linewidth]{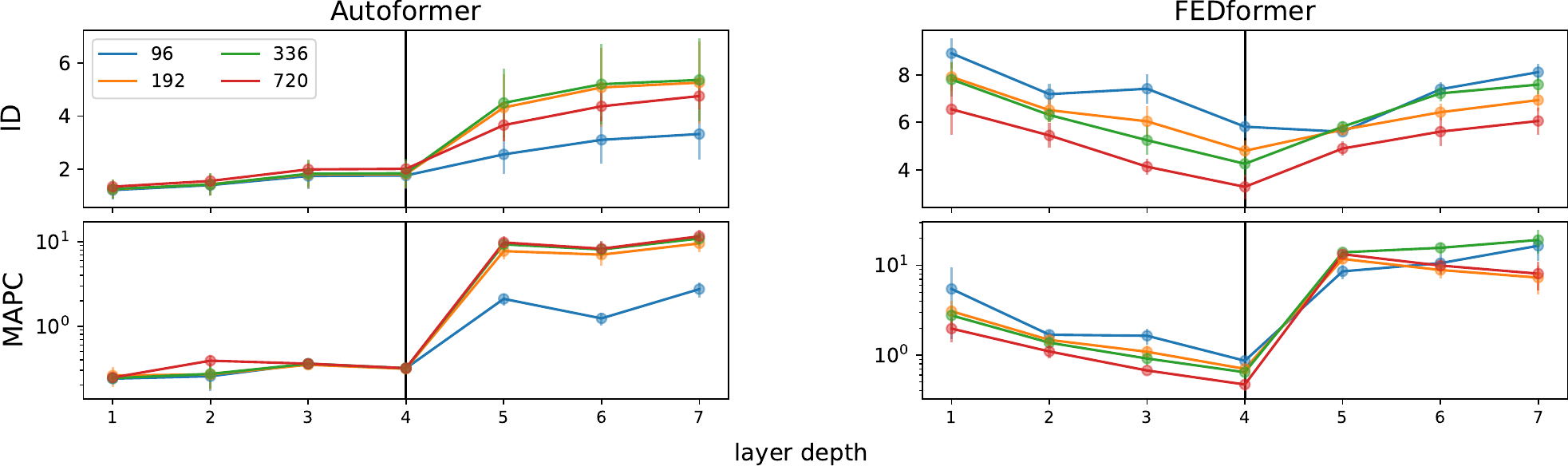}
    \vspace{-3mm}
    \caption{\textbf{Intrinsic dimension and mean absolute principal curvature along the layers of Autoformer and FEDformer on traffic dataset for multiple forecasting horizons.}  Top) intrinsic dimension. Bottom) mean absolute principal curvature. For each model, both ID and MAPC share a similar profile across different forecasting horizons.}
    \label{fig:id_mapc_traffic}
\end{figure*}

\section{Results}

\subsection{Data manifolds share similar geometric profiles}
\label{subsec:sim_id_mapc}

In our first empirical result, we compute the intrinsic dimension (ID) and mean absolute principal curvature (MAPC) across the layers of Autoformer and FEDformer models on the traffic dataset. In Fig.~\ref{fig:id_mapc_traffic}, we plot the ID (top row) and MAPC (bottom row) for Autoformer (left column) and FEDformer (right column) on multiple forecast horizons$\;=96, 192,336,720$. The $x$-labels refer to the layers we sample, where labels \numrange{1}{4} refer to two sequence decomposition layers per encoder block (and thus four in total), labels \numrange{5}{6} denote the decoder decomposition layers, and label $7$ is the linear output layer, see Fig.~\ref{fig:arch} for the network scheme. Our results indicate that during the encoding phase, the ID and the MAPC are relatively fixed for Autoformer and decrease for FEDformer, and during the decoder module, these values generally increase with depth. Specifically, the ID values change from $\min(\text{ID})=1.2$ to $\max(\text{ID})=8.1$, showing a relatively small variation across layers. In comparison, the mean absolute principal curvature values present a larger deviation as they range from $\min(\text{MAPC})=0.2$ to $\max(\text{MAPC})=19.2$.

\begin{figure*}[t]
    \centering
    \includegraphics[width=1\linewidth]{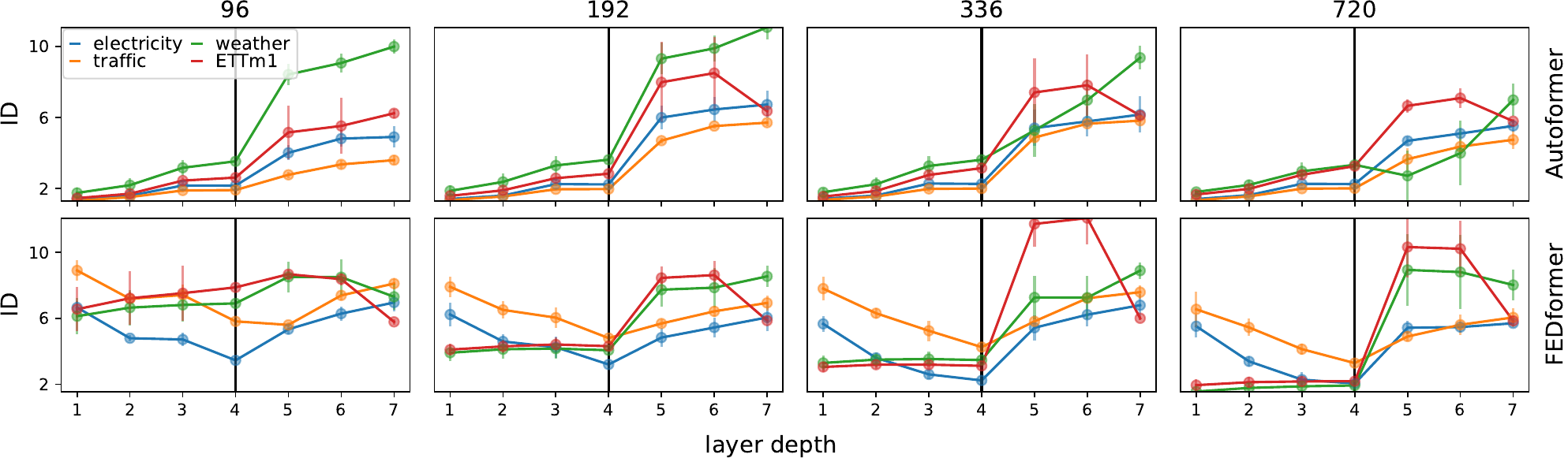}
    \caption{\textbf{ID profiles across layers of Autoformer and FEDformer on electricity, traffic, weather and ETTm1 datasets for multiple forecasting horizons.} Each panel includes ID profiles per dataset, for several horizons (left to right) and architectures (top to bottom).}
    \label{fig:id_auto_fed_mult_datasets}
\end{figure*}

Remarkably, it can be observed from Fig.~\ref{fig:id_mapc_traffic} that both Autoformer and FEDformer feature similar ID and MAPC profiles in terms of values. Further, a strong similarity in trend can be viewed across different forecast horizons per method. Moreover, Autoformer and FEDformer differ during the encoding phase (layers \numrange{1}{4}), but match quite well during the decoding and output phases (layers \numrange{5}{7}). Our intrinsic dimension estimations stand in contrast to existing results on classification tasks with CNN and transformer architectures that observed a ``hunchback'' ID profile~\citep{ansuini2019intrinsic, valeriani2023geometry}. That is, prior work found the intrinsic dimension to increase significantly at the first few layers, and then, it presented a sharp decrease with depth. This qualitative difference in ID can be attributed to the differences between classification models as in~\citet{ansuini2019intrinsic} in comparison to regression TSF networks we consider here. In particular, deep classification neural networks essentially recover the related low-dimensional manifold, to facilitate a linear separation of classes~\citep{cohen2020separability}, and thus one may expect a low ID toward the final layers of the network. On the other hand, forecast regression models aim to encode the statistical distribution of input data, to facilitate forecast of the horizon windows and it is typically of a higher dimension due to spurious variations. In particular, regression TSF models are expected to learn an underlying low-dimensional and simple representation while encoding. Then, a more complex manifold that better reflects the properties of the data is learned during decoding. Importantly, while our intrinsic dimension profiles do not exhibit the ``hunchback'' shape identified in~\citet{ansuini2019intrinsic}, our estimated ID $d$ is significantly smaller than the extrinsic dimension $D=512$, in correspondence with existing work. Finally, our MAPC profiles attain a ``step-like'' appearance, similar to the results in~\citet{kaufman2023data}, where they identify a sharp increase in curvature in the final layer, and we observe such a jump in the decoder.

\begin{figure*}[t]
    \centering
    \includegraphics[width=1\linewidth]{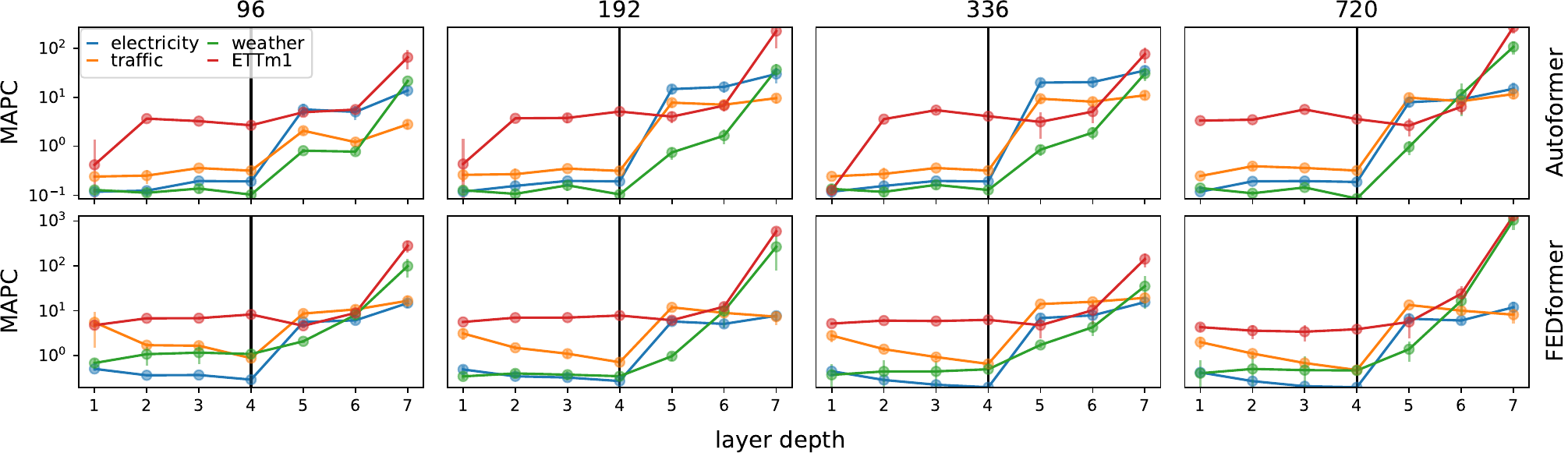}
    \caption{\textbf{MAPC profiles across layers of Autoformer and FEDformer on electricity, traffic, weather and ETTm1 for multiple horizons.} Each panel includes MAPC profiles per dataset, for several horizons (left to right) and architectures (top to bottom).}
    \label{fig:mapc_auto_fed_mult_datasets}
\end{figure*}

To extend our analysis, we present in Fig.~\ref{fig:id_auto_fed_mult_datasets} and \ref{fig:mapc_auto_fed_mult_datasets} the ID and MAPC profiles, respectively, for Autoformer (top) and FEDformer (bottom) for several horizons using multiple different datasets. For all Autoformer configurations, the IDs in Fig.~\ref{fig:id_auto_fed_mult_datasets} generally increase with depth, and the IDs of FEDformer present a ``v''-shape for electricity and traffic and a ``step''-like behavior for weather and ETTm1. Interestingly, ETTm1 (and other ETT* datasets, please see Fig.~\ref{fig:id_ett}) shows a hunchback trend, however, the drop of ID in the final layer is due to ETT* datasets consisting of a total of seven features, and thus we do not consider this behavior to be characteristic to the network. As in Fig.~\ref{fig:id_mapc_traffic} and existing work~\citep{ansuini2019intrinsic}, the intrinsic dimension $d$ is much lower than its extrinsic counterpart $D$. Our MAPC results in Fig.~\ref{fig:mapc_auto_fed_mult_datasets} indicate a shared step-like behavior in general for all models, horizons, and datasets, where the main difference is where the curvature increase occurs. For electricity and traffic, we observe a sharp increase at the beginning of the decoder block, whereas for weather and ETTm1, the increase often appears at the final layer. Additionally, the maximal curvature values for weather and ETTm1 tend to be higher than those of electricity and traffic. Overall, our results suggest that weather and ETTm1 are associated with manifolds whose geometric features match. This observation can be justified by the known correlation between electricity transformer temperature (ETT) and climate change~\citep{hashmi2013effect, gao2018potential}. Similarly, electricity consumption (electricity) and road occupancy (traffic) attain a shared behavior that may be explained due to the strong seasonality component in these datasets~\citep{zeng2023transformers}.

\begin{figure*}[!hb]
    \centering
    \begin{overpic}[width=1\linewidth]{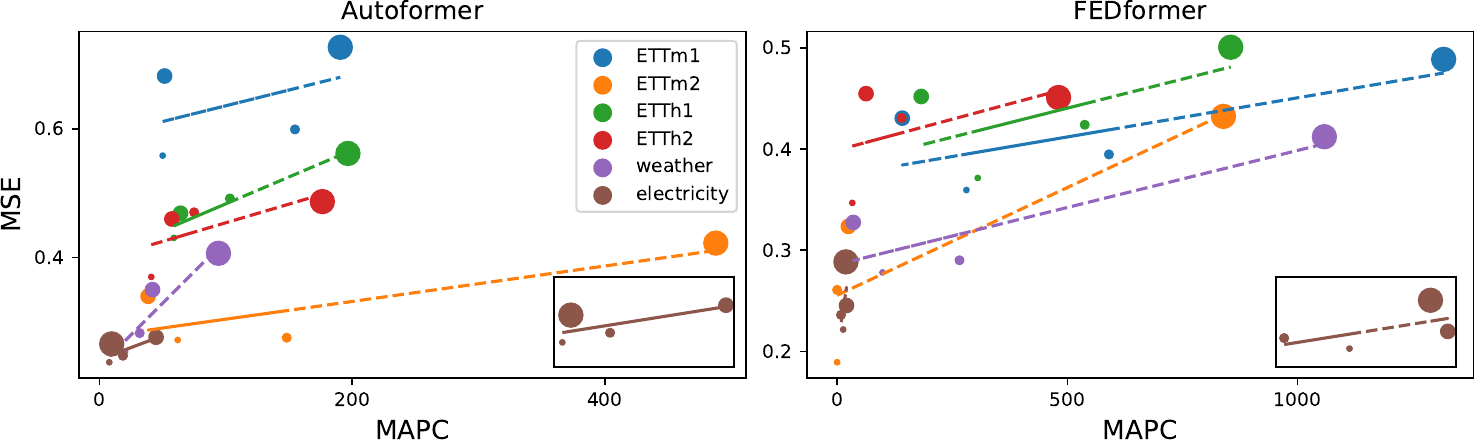}

    \end{overpic}
    \caption{\textbf{MAPC is correlated with model performance.} Each color represents a different dataset while the size of the dot is determined by the forecast horizon (longer horizon results in a larger dot). The test mean squared error is proportional to the MAPC on multiple datasets.} 
    \label{fig:id_vs_mse}
\end{figure*}

\subsection{Final MAPC is correlated with performance}
\label{subsec:mapc_corr}

We further investigate whether geometric properties of the learned manifold are related to inherent features of the model. For instance, previous works find a strong correlation between the ID~\citep{ansuini2019intrinsic} and MAPC~\citep{kaufman2023data} with model performance and generalization. Specifically, the intrinsic dimension in the last hidden layer is correlated with the top-$5$ score on image classification, i.e., lower ID is associated with lower error. Similarly, large normalized MAPC gap between the penultimate and final layers of CNNs is related to high classification accuracy. These correlations are important as they allow developers and practitioners to evaluate and compare deep neural networks based on statistics obtained directly from the train set. This is crucial in scenarios where, e.g., the test set is unavailable during model design.

We show in Fig.~\ref{fig:id_vs_mse} plots of the test mean squared error ($e_\text{MSE}$) vs. the MAPC in the final layer of Autoformer and FEDformer models trained on ETTm1, ETTm2, ETTh1, ETTh2, weather, and electricity. For each dataset, we plot four colored circles corresponding to the four different horizons, where each circle is scaled proportionally to the horizon length. The dashed graphs are generated by plotting the $e_\text{MSE}$ with respect to the MAPC, representing the best linear fit for each dataset. Due to different scales, an inlay of the electricity dataset is added to the bottom right corner of each architecture. We find a positive slope in all Autoformer and FEDformer models with an average correlation coefficient of $0.76$ and $0.7$, respectively (see full results in Tab.~\ref{tab:mapc_corr}). In all cases, we observe a correlation between the test MSE and final MAPC, namely, the model performs \emph{better} as curvature \emph{decreases}, as shown in Fig.~\ref{fig:id_vs_mse}. 

As in Sec.~\ref{subsec:sim_id_mapc}, we identify different characteristics for TSF models with respect to classification neural networks. While popular CNNs show better performance when the MAPC gap is high~\citep{kaufman2023data}, we report an opposite trend, namely, better models are associated with a \emph{lower} MAPC. We believe this behavior may be attributed to classification networks requiring final high curvatures to properly classify the statistical distribution of the input information and its large variance. In addition, we note the relatively flat slope profiles presented across all datasets and architectures. Essentially, these results indicate that while the manifolds become more complex in terms of curvature, transformers yield a relatively fixed MSE, regardless of the underlying MAPC. Thus, our results may hint that Autoformer and FEDformer are \emph{not} expressive enough. Indeed, the datasets we consider consist of many features (on the range of hundreds for electricity and traffic). Therefore, while TSF approaches may need highly expressive networks to model these datasets due to their complex statistics, it might be that current approaches can not achieve better representations in these cases, and they tend to get ``stuck'' on a local minimum. We hypothesize that significantly deeper and more expressive TSF models as in~\citet{nie2023time} and is common in classification~\citep{he2016deep} may yield double descent forecasting architectures~\citep{belkin2019reconciling}.

\begin{table}[t]
    \caption{Correlation coefficients of test MSE vs. final MAPC for several datasets and architectures.}
    \label{tab:mapc_corr}
    \vskip 0.1in
    \centering
    \begin{tabular}{|c|c|c|c|c|c|c|}
        \toprule
        & ETTm1 & ETTm2 & ETTh1 & ETTh2 & Weather & Electricity \\
        \midrule
        Autoformer & 0.46 & 0.81 & 0.97 & 0.67 & 0.93 & 0.71 \\
        FEDformer & 0.74 & 0.86 & 0.63 & 0.50 & 0.58 & 0.70 \\
        \bottomrule
    \end{tabular}
\end{table}

\vspace{-10pt}

\subsection{Manifold dynamics during training}

\begin{figure*}[b!]
    \centering
    \includegraphics[width=1\linewidth]{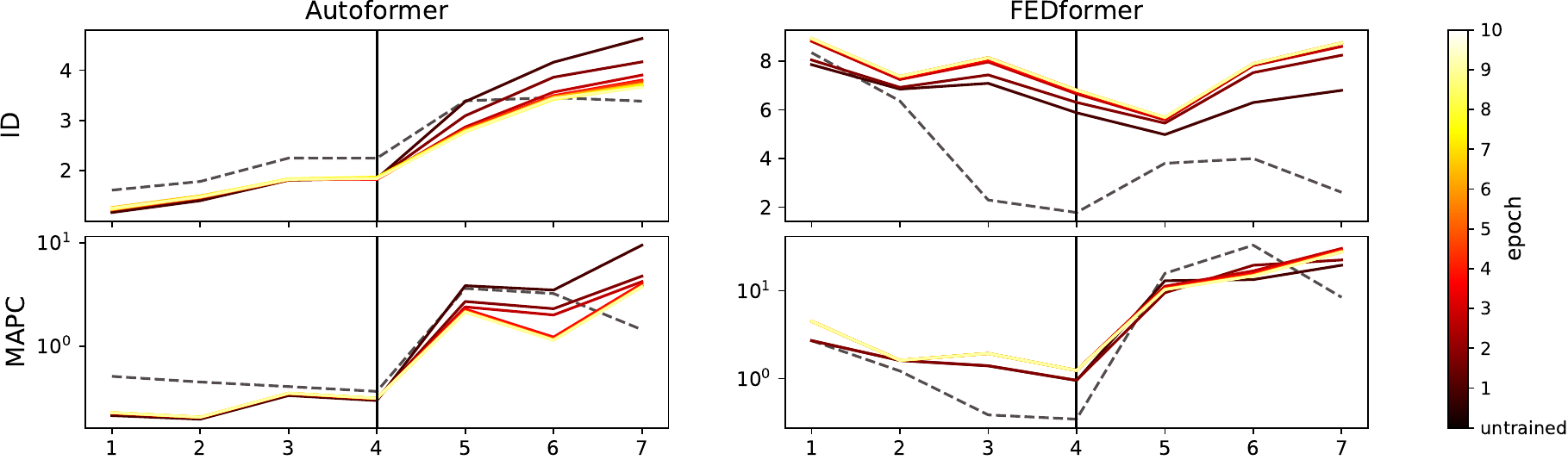}
    \caption{\textbf{Training dynamics of the ID and MAPC on traffic dataset.} The plot shows how the ID and MAPC change during training, colored by the training epoch.}
    \label{fig:auto_fed_train_dynamics}
\end{figure*}

Our analysis above focuses on fully trained deep neural networks and the geometric properties of the learned manifolds. In addition, we also investigate below the evolution of manifolds during training and their structure at initialization. In prior works, \citet{ansuini2019intrinsic} observed that randomly initialized architectures exhibit a constant ID profile, and further, there is an opposite trend in ID in intermediate layers vs. final layers of convolutional neural networks during training. \citet{kaufman2023data} find that untrained models have different MAPC profiles than trained networks, and they observe that the normalized MAPC gap consistently increases with model performance during training. Moreover, ID and MAPC profiles converge consistently to their final configuration as training proceeds. Motivated by their analysis, we study the general convergence and trend of ID and MAPC of TSF models during training evolution.

We show in Fig.~\ref{fig:auto_fed_train_dynamics} the ID and MAPC profiles for Autoformer and FEDformer during training, where each plot is colored by its sampling epoch using the \code{hot} colormap. First, the untrained ID and MAPC profiles (dashed black) are somewhat random in comparison to the other geometric profiles. Second, the overall convergence to the final behavior is extremely fast, requiring approximately five epochs to converge in all the configurations which is consistent with the results of~\citet{bonheme2022fondue} where they show that the ID does not change much after the first epoch. Moreover, the encoder in the Autoformer converges within two epochs, whereas the FEDformer model needs more epochs for the encoder to converge. Third, the decoder shows a slower convergence for both methods, suggesting that ``most'' learning takes place in the decoder component of transformer-based forecasting models. The previous observation aligns with the works of~\citet{bonheme2023good,raghu2017svcca}, showing that representations of layers closer to the input tend to stabilize quicker. More specifically, \citet{bonheme2023good} show that encoders’ representations are generic while decoders’ are specific, resulting in a slight change of the encoders’ representations during training. Finally, except for the untrained profiles, the ID and MAPC curves during training are generally similar across different epochs. The latter observation may mean that Autoformer and FEDformer mainly perform fine-tuning training as their underlying manifolds do not change much during training.

\subsection{Distribution of principal curvatures}

\begin{figure*}[b!]
    \centering
    \includegraphics[width=1\linewidth]{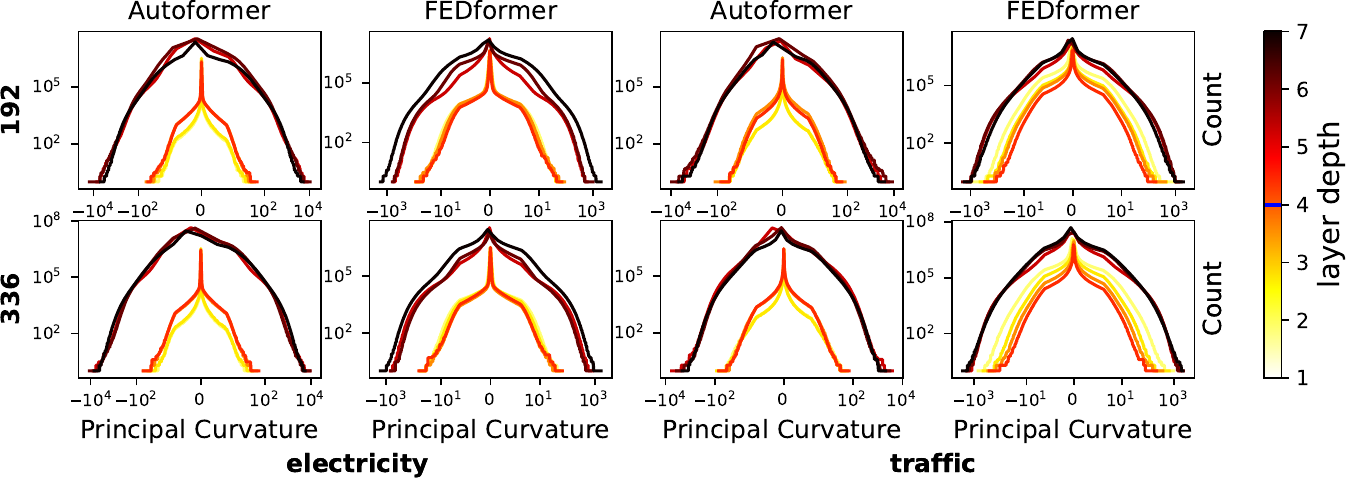}
    \caption{\textbf{Principal curvature distribution of Autoformer and FEDformer.} Each plot shows the histogram profiles of principal curvatures per layer, colored by their relative depth.}
    \label{fig:auto_fed_pc_hist}
\end{figure*}

The CAML algorithm~\citep{li2018curvature} we employ for estimating the principal curvatures produces $d(D-d)$ values per point, yielding a massive amount of curvature information for analysis. Following~\citet{kaufman2023data}, we compute and plot in Fig.~\ref{fig:auto_fed_pc_hist} the distribution of principal curvatures for every layer, shown as a smooth histogram for Autoformer and FEDformer models on electricity and traffic datasets with horizons$\;=192, 336$. The histogram plots are colored by the network depth using the \code{hot} colormap. These distributions strengthen our analysis in Sec.~\ref{subsec:sim_id_mapc} where we observe a ``step''-like pattern in MAPC, where the sharp jump in curvature occurs at the beginning of the decoder. Indeed, the curves in Fig.~\ref{fig:auto_fed_pc_hist} related to layers \numrange{1}{4} span a smaller range in comparison to the curves in layers \numrange{5}{7}. Further, the histograms show that the distribution of curvature is relatively fixed across the encoder blocks, and similarly, a different but rather fixed profile appears in the decoder.

\section{Discussion}

Deep neural networks are composed of several computation layers. Each layer receives inputs from a preceding layer, it applies a (nonlinear) transformation, and it feeds the outputs to a subsequent layer. The overarching theme in this work is the investigation of data representations arising during the computation of deep models. To study these latent representations, we adopt a ubiquitous ansatz, commonly known as the \emph{manifold hypothesis}~\citep{coifman2006diffusion}: We assume that while such data may be given in a high-dimensional and complex format, it lies on or next to a low-dimensional manifold. The implications of this inductive bias are paramount; manifolds are rich mathematical objects that are actively studied in theory~\citep{lee2006riemannian} and practice~\citep{chaudhry2018riemannian}, allowing one to harness the abundant classical tools and recent developments to study deep representations.

Our study aligns with the line of works that aims at better understanding the inner mechanisms of deep neural networks. Indeed, while modern machine learning has been dominating many scientific and engineering disciplines since the appearance of AlexNet~\citep{krizhevsky2012imagenet}, neural net architectures are still considered not well understood by many. In this context, the manifold ansatz is instrumental---existing analysis works investigate geometric features of latent manifolds and their relation to the underlying task and model performance. For instance, \citet{ansuini2019intrinsic} compute the intrinsic dimension of popular convolutional neural networks. Following their work, \citet{kaufman2023data} estimate the mean absolute principal curvatures. However, while CNNs are relatively studied from the manifold viewpoint, impactful sequential transformer models \citep{vaswani2017attention}, received less attention~\citep{valeriani2023geometry}. The lack of analysis is even more noticeable for the recent state-of-the-art transformer-based \emph{time series forecasting} works, e.g.,~\citet{wu2021autoformer}. The main objective of our work is to help bridge this gap and study deep forecasting models trained on common challenging datasets from a manifold learning viewpoint.

We compute the ID and MAPC of data representations from several different deep architectures, forecasting tasks and datasets. To this end, we employ differentiable tools~\citep{facco2017estimating, li2018curvature} that produce a single scalar ID and many principle curvatures combined to a single scalar MAPC, per manifold. Our results raise several intriguing observations, many of them are in correspondence with existing work. First, the ID is much smaller than the extrinsic dimension, reflecting that learned manifolds are indeed low-dimensional. Second, the ID and MAPC profiles across layers are similar for many different architectures, tasks, and datasets. In particular, we identify two phases, where in the encoder, ID and MAPC are decreasing or stay fixed, and in the decoder, both geometric features increase with depth. Third, the MAPC in the final layer is strongly correlated with model performance, presenting a correlation, i.e., error is lower when MAPC is lower. Fourth, we observe that related but different datasets attain similar manifolds, whereas unrelated datasets are associated to manifolds with different characteristics. Finally, untrained models present random ID and MAPC profiles that converge to their final configuration within a few epochs.

Our analysis and observations lie at the heart of the differences between classification and regression tasks; a research avenue that only recently had started to be addressed more frequently~\citep{muthukumar2021classification,yao2022c}. Our results indicate a fundamental difference between image classification and time series forecasting regression models: while the former networks shrink the ID significantly to extract a meaningful representation that is amenable for linear separation, TSF models behave differently. Indeed, the ID generally increases with depth, perhaps to properly capture the large variance of the input domain where regression networks predict. Moreover, while a high MAPC gap seems to be important for classification, we find an opposite trend in TSF regression problems. In conclusion, we believe that our work sets the stage for a deeper investigation of classification vs. regression from a manifold learning and other perspectives. We believe that fundamental advancements on this front will lead to powerful machine learning models, better suited for solving the task at hand.

\clearpage
\section*{Acknowledgements}
\label{Sec:acknowledgement}
This research was partially supported by the Lynn and William Frankel Center of the Computer Science Department, Ben-Gurion University of the Negev, an ISF grant 668/21, an ISF equipment grant, and by the Israeli Council for Higher Education (CHE) via the Data Science Research Center, Ben-Gurion University of the Negev, Israel.

\bibliography{main}

\begin{thebibliography}{71}
\providecommand{\natexlab}[1]{#1}
\providecommand{\url}[1]{\texttt{#1}}
\expandafter\ifx\csname urlstyle\endcsname\relax
  \providecommand{\doi}[1]{doi: #1}\else
  \providecommand{\doi}{doi: \begingroup \urlstyle{rm}\Url}\fi

\bibitem[Anderson \& Kendall(1976)Anderson and Kendall]{Anderson1976TimeSeries2E}
Oliver~D. Anderson and M.~G. Kendall.
\newblock Time-series. 2nd edn.
\newblock \emph{The Statistician}, 25:\penalty0 308, 1976.
\newblock URL \url{https://api.semanticscholar.org/CorpusID:134001785}.

\bibitem[Ansuini et~al.(2019)Ansuini, Laio, Macke, and Zoccolan]{ansuini2019intrinsic}
Alessio Ansuini, Alessandro Laio, Jakob~H Macke, and Davide Zoccolan.
\newblock Intrinsic dimension of data representations in deep neural networks.
\newblock \emph{Advances in Neural Information Processing Systems}, 32, 2019.

\bibitem[Bac et~al.(2021)Bac, Mirkes, Gorban, Tyukin, and Zinovyev]{bac2021scikit}
Jonathan Bac, Evgeny~M Mirkes, Alexander~N Gorban, Ivan Tyukin, and Andrei Zinovyev.
\newblock Scikit-dimension: a python package for intrinsic dimension estimation.
\newblock \emph{Entropy}, 23\penalty0 (10):\penalty0 1368, 2021.

\bibitem[Bahdanau et~al.(2015)Bahdanau, Cho, and Bengio]{bahdanau2014neural}
Dzmitry Bahdanau, Kyunghyun Cho, and Yoshua Bengio.
\newblock Neural machine translation by jointly learning to align and translate.
\newblock In \emph{3rd International Conference on Learning Representations, {ICLR}}, 2015.

\bibitem[Belkin et~al.(2019)Belkin, Hsu, Ma, and Mandal]{belkin2019reconciling}
Mikhail Belkin, Daniel Hsu, Siyuan Ma, and Soumik Mandal.
\newblock Reconciling modern machine-learning practice and the classical bias--variance trade-off.
\newblock \emph{Proceedings of the National Academy of Sciences}, 116\penalty0 (32):\penalty0 15849--15854, 2019.

\bibitem[Berman et~al.(2023)Berman, Naiman, and Azencot]{berman2023multifactor}
Nimrod Berman, Ilan Naiman, and Omri Azencot.
\newblock Multifactor sequential disentanglement via structured {Koopman} autoencoders.
\newblock In \emph{The Eleventh International Conference on Learning Representations, {ICLR}}, 2023.

\bibitem[Berman et~al.(2024)Berman, Naiman, Arbiv, Fadlon, and Azencot]{berman2024sequential}
Nimrod Berman, Ilan Naiman, Idan Arbiv, Gal Fadlon, and Omri Azencot.
\newblock Sequential disentanglement by extracting static information from {A} single sequence element.
\newblock In \emph{Forty-first International Conference on Machine Learning, {ICML}}, 2024.

\bibitem[Bonheme \& Grzes(2022)Bonheme and Grzes]{bonheme2022fondue}
Lisa Bonheme and Marek Grzes.
\newblock Fondue: an algorithm to find the optimal dimensionality of the latent representations of variational autoencoders.
\newblock \emph{arXiv preprint arXiv:2209.12806}, 2022.

\bibitem[Bonheme \& Grzes(2023)Bonheme and Grzes]{bonheme2023good}
Lisa Bonheme and Marek Grzes.
\newblock How good are variational autoencoders at transfer learning?
\newblock \emph{arXiv preprint arXiv:2304.10767}, 2023.

\bibitem[Brahma et~al.(2015)Brahma, Wu, and She]{brahma2015deep}
Pratik~Prabhanjan Brahma, Dapeng Wu, and Yiyuan She.
\newblock Why deep learning works: A manifold disentanglement perspective.
\newblock \emph{IEEE transactions on neural networks and learning systems}, 27\penalty0 (10):\penalty0 1997--2008, 2015.

\bibitem[Caron et~al.(2021)Caron, Touvron, Misra, J{\'e}gou, Mairal, Bojanowski, and Joulin]{caron2021emerging}
Mathilde Caron, Hugo Touvron, Ishan Misra, Herv{\'e} J{\'e}gou, Julien Mairal, Piotr Bojanowski, and Armand Joulin.
\newblock Emerging properties in self-supervised vision transformers.
\newblock In \emph{Proceedings of the IEEE/CVF international conference on computer vision}, pp.\  9650--9660, 2021.

\bibitem[Challu et~al.(2022)Challu, Olivares, Oreshkin, Garza, Mergenthaler, and Dubrawski]{challu2022n}
Cristian Challu, Kin~G Olivares, Boris~N Oreshkin, Federico Garza, Max Mergenthaler, and Artur Dubrawski.
\newblock {N-HiTS}: Neural hierarchical interpolation for time series forecasting.
\newblock \emph{arXiv preprint arXiv:2201.12886}, 2022.

\bibitem[Chaudhry et~al.(2018)Chaudhry, Dokania, Ajanthan, and Torr]{chaudhry2018riemannian}
Arslan Chaudhry, Puneet~K Dokania, Thalaiyasingam Ajanthan, and Philip~HS Torr.
\newblock Riemannian walk for incremental learning: Understanding forgetting and intransigence.
\newblock In \emph{Proceedings of the European conference on computer vision (ECCV)}, pp.\  532--547, 2018.

\bibitem[Chefer et~al.(2021)Chefer, Gur, and Wolf]{chefer2021transformer}
Hila Chefer, Shir Gur, and Lior Wolf.
\newblock Transformer interpretability beyond attention visualization.
\newblock In \emph{Proceedings of the IEEE/CVF conference on computer vision and pattern recognition}, pp.\  782--791, 2021.

\bibitem[Clark et~al.(2019)Clark, Khandelwal, Levy, and Manning]{clark2019does}
Kevin Clark, Urvashi Khandelwal, Omer Levy, and Christopher~D Manning.
\newblock What does bert look at? an analysis of bert’s attention.
\newblock In \emph{Proceedings of the 2019 ACL Workshop BlackboxNLP: Analyzing and Interpreting Neural Networks for NLP}. Association for Computational Linguistics, 2019.

\bibitem[Cleveland et~al.(1990)Cleveland, Cleveland, McRae, Terpenning, et~al.]{cleveland1990stl}
Robert~B Cleveland, William~S Cleveland, Jean~E McRae, Irma Terpenning, et~al.
\newblock Stl: A seasonal-trend decomposition.
\newblock \emph{J. off. Stat}, 6\penalty0 (1):\penalty0 3--73, 1990.

\bibitem[Cohen et~al.(2020)Cohen, Chung, Lee, and Sompolinsky]{cohen2020separability}
Uri Cohen, SueYeon Chung, Daniel~D Lee, and Haim Sompolinsky.
\newblock Separability and geometry of object manifolds in deep neural networks.
\newblock \emph{Nature communications}, 11\penalty0 (1):\penalty0 746, 2020.

\bibitem[Coifman \& Lafon(2006)Coifman and Lafon]{coifman2006diffusion}
Ronald~R Coifman and St{\'e}phane Lafon.
\newblock Diffusion maps.
\newblock \emph{Applied and computational harmonic analysis}, 21\penalty0 (1):\penalty0 5--30, 2006.

\bibitem[Devlin et~al.(2019)Devlin, Chang, Lee, and Toutanova]{devlin2019bert}
Jacob Devlin, Ming{-}Wei Chang, Kenton Lee, and Kristina Toutanova.
\newblock {BERT:} pre-training of deep bidirectional transformers for language understanding.
\newblock In \emph{Proceedings of the 2019 Conference of the North American Chapter of the Association for Computational Linguistics: Human Language Technologies, {NAACL-HLT}}. Association for Computational Linguistics, 2019.

\bibitem[Doimo et~al.(2020)Doimo, Glielmo, Ansuini, and Laio]{doimo2020hierarchical}
Diego Doimo, Aldo Glielmo, Alessio Ansuini, and Alessandro Laio.
\newblock Hierarchical nucleation in deep neural networks.
\newblock \emph{Advances in Neural Information Processing Systems}, 33:\penalty0 7526--7536, 2020.

\bibitem[Dosovitskiy et~al.(2020)Dosovitskiy, Beyer, Kolesnikov, Weissenborn, Zhai, Unterthiner, Dehghani, Minderer, Heigold, Gelly, et~al.]{dosovitskiy2020image}
Alexey Dosovitskiy, Lucas Beyer, Alexander Kolesnikov, Dirk Weissenborn, Xiaohua Zhai, Thomas Unterthiner, Mostafa Dehghani, Matthias Minderer, Georg Heigold, Sylvain Gelly, et~al.
\newblock An image is worth 16x16 words: Transformers for image recognition at scale.
\newblock In \emph{International Conference on Learning Representations}, 2020.

\bibitem[Facco et~al.(2017)Facco, d’Errico, Rodriguez, and Laio]{facco2017estimating}
Elena Facco, Maria d’Errico, Alex Rodriguez, and Alessandro Laio.
\newblock Estimating the intrinsic dimension of datasets by a minimal neighborhood information.
\newblock \emph{Scientific reports}, 7\penalty0 (1):\penalty0 12140, 2017.

\bibitem[Gao et~al.(2018)Gao, Schlosser, and Morgan]{gao2018potential}
Xiang Gao, C~Adam Schlosser, and Eric~R Morgan.
\newblock Potential impacts of climate warming and increased summer heat stress on the electric grid: a case study for a large power transformer (lpt) in the northeast united states.
\newblock \emph{Climatic change}, 147:\penalty0 107--118, 2018.

\bibitem[Goodfellow et~al.(2016)Goodfellow, Bengio, and Courville]{goodfellow2016deep}
Ian Goodfellow, Yoshua Bengio, and Aaron Courville.
\newblock \emph{Deep learning}.
\newblock MIT press, 2016.

\bibitem[Graves et~al.(2013)Graves, Mohamed, and Hinton]{graves2013speech}
Alex Graves, Abdel-rahman Mohamed, and Geoffrey Hinton.
\newblock Speech recognition with deep recurrent neural networks.
\newblock In \emph{2013 IEEE international conference on acoustics, speech and signal processing}, pp.\  6645--6649. Ieee, 2013.

\bibitem[Han et~al.(2018)Han, Feng, Chen, Xu, and Qiu]{han2018structured}
Min Han, Shoubo Feng, CL~Philip Chen, Meiling Xu, and Tie Qiu.
\newblock Structured manifold broad learning system: A manifold perspective for large-scale chaotic time series analysis and prediction.
\newblock \emph{IEEE Transactions on Knowledge and Data Engineering}, 31\penalty0 (9):\penalty0 1809--1821, 2018.

\bibitem[Hashimoto et~al.(2016)Hashimoto, Alvarez-Melis, and Jaakkola]{hashimoto2016word}
Tatsunori~B Hashimoto, David Alvarez-Melis, and Tommi~S Jaakkola.
\newblock Word embeddings as metric recovery in semantic spaces.
\newblock \emph{Transactions of the Association for Computational Linguistics}, 4:\penalty0 273--286, 2016.

\bibitem[Hashmi et~al.(2013)Hashmi, Lehtonen, Seppo, et~al.]{hashmi2013effect}
Murtaza Hashmi, Matti Lehtonen, H~Seppo, et~al.
\newblock Effect of climate change on transformers loading conditions in the future smart grid environment.
\newblock \emph{Open Journal of Applied Sciences}, 3\penalty0 (02):\penalty0 24, 2013.

\bibitem[Hauser \& Ray(2017)Hauser and Ray]{hauser2017principles}
Michael Hauser and Asok Ray.
\newblock Principles of riemannian geometry in neural networks.
\newblock \emph{Advances in neural information processing systems}, 30, 2017.

\bibitem[He et~al.(2016)He, Zhang, Ren, and Sun]{he2016deep}
Kaiming He, Xiangyu Zhang, Shaoqing Ren, and Jian Sun.
\newblock Deep residual learning for image recognition.
\newblock In \emph{Proceedings of the IEEE conference on computer vision and pattern recognition}, pp.\  770--778, 2016.

\bibitem[Jacot et~al.(2018)Jacot, Gabriel, and Hongler]{jacot2018neural}
Arthur Jacot, Franck Gabriel, and Cl{\'e}ment Hongler.
\newblock Neural tangent kernel: Convergence and generalization in neural networks.
\newblock \emph{Advances in neural information processing systems}, 31, 2018.

\bibitem[Karpathy et~al.(2015)Karpathy, Johnson, and Fei-Fei]{karpathy2015visualizing}
Andrej Karpathy, Justin Johnson, and Li~Fei-Fei.
\newblock Visualizing and understanding recurrent networks.
\newblock \emph{arXiv preprint arXiv:1506.02078}, 2015.

\bibitem[Kaufman \& Azencot(2023)Kaufman and Azencot]{kaufman2023data}
Ilya Kaufman and Omri Azencot.
\newblock Data representations' study of latent image manifolds.
\newblock In \emph{International Conference on Machine Learning, {ICML}}, 2023.

\bibitem[Kaufman \& Azencot(2024)Kaufman and Azencot]{kaufman2024first}
Ilya Kaufman and Omri Azencot.
\newblock First-order manifold data augmentation for regression learning.
\newblock In \emph{Forty-first International Conference on Machine Learning, {ICML}}, 2024.

\bibitem[Krizhevsky et~al.(2012)Krizhevsky, Sutskever, and Hinton]{krizhevsky2012imagenet}
Alex Krizhevsky, Ilya Sutskever, and Geoffrey~E Hinton.
\newblock Imagenet classification with deep convolutional neural networks.
\newblock \emph{Advances in neural information processing systems}, 25, 2012.

\bibitem[Lee(2006)]{lee2006riemannian}
John~M Lee.
\newblock \emph{Riemannian manifolds: an introduction to curvature}, volume 176.
\newblock Springer Science \& Business Media, 2006.

\bibitem[Li et~al.(2021)Li, Wei, and Dai]{li2021combination}
Jinghua Li, Shanyang Wei, and Wei Dai.
\newblock Combination of manifold learning and deep learning algorithms for mid-term electrical load forecasting.
\newblock \emph{IEEE Transactions on Neural Networks and Learning Systems}, 2021.

\bibitem[Li(2018)]{li2018curvature}
Yangyang Li.
\newblock Curvature-aware manifold learning.
\newblock \emph{Pattern Recognition}, 83:\penalty0 273--286, 2018.

\bibitem[Li \& Liang(2018)Li and Liang]{li2018learning}
Yuanzhi Li and Yingyu Liang.
\newblock Learning overparameterized neural networks via stochastic gradient descent on structured data.
\newblock \emph{Advances in neural information processing systems}, 31, 2018.

\bibitem[Lin et~al.(2006)Lin, Liu, Yang, Ahuja, and Levinson]{lin2006learning}
Ruei{-}Sung Lin, Che{-}Bin Liu, Ming{-}Hsuan Yang, Narendra Ahuja, and Stephen~E. Levinson.
\newblock Learning nonlinear manifolds from time series.
\newblock In \emph{Computer Vision - {ECCV}}, 2006.

\bibitem[Liu et~al.(2021)Liu, Yu, Liao, Li, Lin, Liu, and Dustdar]{liu2021pyraformer}
Shizhan Liu, Hang Yu, Cong Liao, Jianguo Li, Weiyao Lin, Alex~X Liu, and Schahram Dustdar.
\newblock Pyraformer: Low-complexity pyramidal attention for long-range time series modeling and forecasting.
\newblock In \emph{International conference on learning representations}, 2021.

\bibitem[Muthukumar et~al.(2021)Muthukumar, Narang, Subramanian, Belkin, Hsu, and Sahai]{muthukumar2021classification}
Vidya Muthukumar, Adhyyan Narang, Vignesh Subramanian, Mikhail Belkin, Daniel Hsu, and Anant Sahai.
\newblock Classification vs regression in overparameterized regimes: Does the loss function matter?
\newblock \emph{The Journal of Machine Learning Research}, 22\penalty0 (1):\penalty0 10104--10172, 2021.

\bibitem[Naiman \& Azencot(2023)Naiman and Azencot]{naiman2023operator}
Ilan Naiman and Omri Azencot.
\newblock An operator theoretic approach for analyzing sequence neural networks.
\newblock In \emph{Thirty-Seventh {AAAI} Conference on Artificial Intelligence, {AAAI}}, pp.\  9268--9276. {AAAI} Press, 2023.

\bibitem[Naiman et~al.(2023)Naiman, Berman, and Azencot]{naiman2023sample}
Ilan Naiman, Nimrod Berman, and Omri Azencot.
\newblock Sample and predict your latent: modality-free sequential disentanglement via contrastive estimation.
\newblock In \emph{International Conference on Machine Learning}, pp.\  25694--25717. PMLR, 2023.

\bibitem[Naiman et~al.(2024)Naiman, Erichson, Ren, Mahoney, and Azencot]{naiman2024generative}
Ilan Naiman, N.~Benjamin Erichson, Pu~Ren, Michael~W. Mahoney, and Omri Azencot.
\newblock Generative modeling of regular and irregular time series data via {Koopman VAE}s.
\newblock In \emph{The Twelfth International Conference on Learning Representations, {ICLR}}, 2024.

\bibitem[Nguyen et~al.(2019)Nguyen, Brun, L{\'e}zoray, and Bougleux]{nguyen2019neural}
Xuan~Son Nguyen, Luc Brun, Olivier L{\'e}zoray, and S{\'e}bastien Bougleux.
\newblock A neural network based on spd manifold learning for skeleton-based hand gesture recognition.
\newblock In \emph{Proceedings of the IEEE/CVF Conference on Computer Vision and Pattern Recognition}, pp.\  12036--12045, 2019.

\bibitem[Nie et~al.(2023)Nie, Nguyen, Sinthong, and Kalagnanam]{nie2023time}
Yuqi Nie, Nam~H. Nguyen, Phanwadee Sinthong, and Jayant Kalagnanam.
\newblock A time series is worth 64 words: Long-term forecasting with transformers.
\newblock In \emph{The Eleventh International Conference on Learning Representations, {ICLR}}, 2023.

\bibitem[Nochumsohn \& Azencot(2024)Nochumsohn and Azencot]{nochumsohn2024data}
Liran Nochumsohn and Omri Azencot.
\newblock Data augmentation policy search for long-term forecasting.
\newblock \emph{arXiv preprint arXiv:2405.00319}, 2024.

\bibitem[O'Reilly et~al.(2017)O'Reilly, Moessner, and Nati]{oReilly2017univariate}
Colin O'Reilly, Klaus Moessner, and Michele Nati.
\newblock Univariate and multivariate time series manifold learning.
\newblock \emph{Knowl. Based Syst.}, 133:\penalty0 1--16, 2017.

\bibitem[Oreshkin et~al.(2020)Oreshkin, Carpov, Chapados, and Bengio]{oreshkin2019n}
Boris~N. Oreshkin, Dmitri Carpov, Nicolas Chapados, and Yoshua Bengio.
\newblock {N-BEATS:} neural basis expansion analysis for interpretable time series forecasting.
\newblock In \emph{8th International Conference on Learning Representations, {ICLR}}, 2020.

\bibitem[Papaioannou et~al.(2022)Papaioannou, Talmon, Kevrekidis, and Siettos]{papaioannou2022time}
Panagiotis~G Papaioannou, Ronen Talmon, Ioannis~G Kevrekidis, and Constantinos Siettos.
\newblock Time-series forecasting using manifold learning, radial basis function interpolation, and geometric harmonics.
\newblock \emph{Chaos: An Interdisciplinary Journal of Nonlinear Science}, 32\penalty0 (8), 2022.

\bibitem[Pennington et~al.(2014)Pennington, Socher, and Manning]{pennington2014glove}
Jeffrey Pennington, Richard Socher, and Christopher~D Manning.
\newblock Glove: Global vectors for word representation.
\newblock In \emph{Proceedings of the 2014 conference on empirical methods in natural language processing (EMNLP)}, pp.\  1532--1543, 2014.

\bibitem[Raghu et~al.(2017)Raghu, Gilmer, Yosinski, and Sohl-Dickstein]{raghu2017svcca}
Maithra Raghu, Justin Gilmer, Jason Yosinski, and Jascha Sohl-Dickstein.
\newblock Svcca: Singular vector canonical correlation analysis for deep learning dynamics and interpretability.
\newblock \emph{Advances in neural information processing systems}, 30, 2017.

\bibitem[Rodrigues et~al.(2018)Rodrigues, Congedo, and Jutten]{rodrigues2018multivariate}
Pedro Luiz~Coelho Rodrigues, Marco Congedo, and Christian Jutten.
\newblock Multivariate time-series analysis via manifold learning.
\newblock In \emph{2018 {IEEE} Statistical Signal Processing Workshop, {SSP}}, pp.\  573--577, 2018.

\bibitem[Rogers et~al.(2021)Rogers, Kovaleva, and Rumshisky]{rogers2021primer}
Anna Rogers, Olga Kovaleva, and Anna Rumshisky.
\newblock A primer in bertology: What we know about how bert works.
\newblock \emph{Transactions of the Association for Computational Linguistics}, 8:\penalty0 842--866, 2021.

\bibitem[Shao et~al.(2018)Shao, Kumar, and Thomas~Fletcher]{shao2018riemannian}
Hang Shao, Abhishek Kumar, and P~Thomas~Fletcher.
\newblock The riemannian geometry of deep generative models.
\newblock In \emph{Proceedings of the IEEE Conference on Computer Vision and Pattern Recognition Workshops}, pp.\  315--323, 2018.

\bibitem[Shnitzer et~al.(2017)Shnitzer, Talmon, and Slotine]{shnitzer2017manifold}
Tal Shnitzer, Ronen Talmon, and Jean{-}Jacques~E. Slotine.
\newblock Manifold learning with contracting observers for data-driven time-series analysis.
\newblock \emph{{IEEE} Trans. Signal Process.}, 65\penalty0 (4):\penalty0 904--918, 2017.

\bibitem[Stephenson et~al.(2021)Stephenson, Padhy, Ganesh, Hui, Tang, and Chung]{stephenson2021geometry}
Cory Stephenson, Suchismita Padhy, Abhinav Ganesh, Yue Hui, Hanlin Tang, and Sue~Yeon Chung.
\newblock On the geometry of generalization and memorization in deep neural networks.
\newblock In \emph{9th International Conference on Learning Representations, ICLR 2021}, 2021.

\bibitem[Tenney et~al.(2019)Tenney, Das, and Pavlick]{tenney2019bert}
Ian Tenney, Dipanjan Das, and Ellie Pavlick.
\newblock Bert rediscovers the classical nlp pipeline.
\newblock In \emph{Proceedings of the 57th Annual Meeting of the Association for Computational Linguistics}, pp.\  4593--4601, 2019.

\bibitem[Tsai et~al.(2019)Tsai, Bai, Yamada, Morency, and Salakhutdinov]{tsai2019transformer}
Yao-Hung~Hubert Tsai, Shaojie Bai, Makoto Yamada, Louis-Philippe Morency, and Ruslan Salakhutdinov.
\newblock Transformer dissection: a unified understanding of transformer's attention via the lens of kernel.
\newblock \emph{arXiv preprint arXiv:1908.11775}, 2019.

\bibitem[Valeriani et~al.(2023)Valeriani, Doimo, Cuturello, Laio, Ansuini, and Cazzaniga]{valeriani2023geometry}
Lucrezia Valeriani, Diego Doimo, Francesca Cuturello, Alessandro Laio, Alessio Ansuini, and Alberto Cazzaniga.
\newblock The geometry of hidden representations of large transformer models.
\newblock \emph{arXiv preprint arXiv:2302.00294}, 2023.

\bibitem[Vaswani et~al.(2017)Vaswani, Shazeer, Parmar, Uszkoreit, Jones, Gomez, Kaiser, and Polosukhin]{vaswani2017attention}
Ashish Vaswani, Noam Shazeer, Niki Parmar, Jakob Uszkoreit, Llion Jones, Aidan~N Gomez, {\L}ukasz Kaiser, and Illia Polosukhin.
\newblock Attention is all you need.
\newblock \emph{Advances in neural information processing systems}, 30, 2017.

\bibitem[Vig(2019)]{vig2019multiscale}
Jesse Vig.
\newblock A multiscale visualization of attention in the transformer model.
\newblock In \emph{Proceedings of the 57th Annual Meeting of the Association for Computational Linguistics: System Demonstrations}, pp.\  37--42, 2019.

\bibitem[Wallace et~al.(2019)Wallace, Tuyls, Wang, Subramanian, Gardner, and Singh]{wallace2019allennlp}
Eric Wallace, Jens Tuyls, Junlin Wang, Sanjay Subramanian, Matt Gardner, and Sameer Singh.
\newblock Allennlp interpret: {A} framework for explaining predictions of {NLP} models.
\newblock In \emph{Proceedings of the 2019 Conference on Empirical Methods in Natural Language Processing and the 9th International Joint Conference on Natural Language Processing, {EMNLP-IJCNLP}}, pp.\  7--12, 2019.

\bibitem[Wu et~al.(2021)Wu, Xu, Wang, and Long]{wu2021autoformer}
Haixu Wu, Jiehui Xu, Jianmin Wang, and Mingsheng Long.
\newblock Autoformer: Decomposition transformers with auto-correlation for long-term series forecasting.
\newblock \emph{Advances in Neural Information Processing Systems}, 34:\penalty0 22419--22430, 2021.

\bibitem[Yao et~al.(2022)Yao, Wang, Zhang, Zou, and Finn]{yao2022c}
Huaxiu Yao, Yiping Wang, Linjun Zhang, James~Y Zou, and Chelsea Finn.
\newblock C-mixup: Improving generalization in regression.
\newblock \emph{Advances in Neural Information Processing Systems}, 35:\penalty0 3361--3376, 2022.

\bibitem[Yu et~al.(2018)Yu, Long, and Hopcroft]{yu2018curvature}
Tao Yu, Huan Long, and John~E Hopcroft.
\newblock Curvature-based comparison of two neural networks.
\newblock In \emph{2018 24th International Conference on Pattern Recognition (ICPR)}, pp.\  441--447. IEEE, 2018.

\bibitem[Zeiler \& Fergus(2014)Zeiler and Fergus]{zeiler2014visualizing}
Matthew~D Zeiler and Rob Fergus.
\newblock Visualizing and understanding convolutional networks.
\newblock In \emph{Computer Vision--ECCV 2014: 13th European Conference, Zurich, Switzerland, September 6-12, 2014, Proceedings, Part I 13}, pp.\  818--833. Springer, 2014.

\bibitem[Zeng et~al.(2023)Zeng, Chen, Zhang, and Xu]{zeng2023transformers}
Ailing Zeng, Muxi Chen, Lei Zhang, and Qiang Xu.
\newblock Are transformers effective for time series forecasting?
\newblock In \emph{Proceedings of the AAAI conference on artificial intelligence}, volume~37, pp.\  11121--11128, 2023.

\bibitem[Zhou et~al.(2021)Zhou, Zhang, Peng, Zhang, Li, Xiong, and Zhang]{zhou2021informer}
Haoyi Zhou, Shanghang Zhang, Jieqi Peng, Shuai Zhang, Jianxin Li, Hui Xiong, and Wancai Zhang.
\newblock Informer: Beyond efficient transformer for long sequence time-series forecasting.
\newblock In \emph{Proceedings of the AAAI conference on artificial intelligence}, volume~35, pp.\  11106--11115, 2021.

\bibitem[Zhou et~al.(2022)Zhou, Ma, Wen, Wang, Sun, and Jin]{zhou2022fedformer}
Tian Zhou, Ziqing Ma, Qingsong Wen, Xue Wang, Liang Sun, and Rong Jin.
\newblock Fedformer: Frequency enhanced decomposed transformer for long-term series forecasting.
\newblock In \emph{International Conference on Machine Learning}, pp.\  27268--27286. PMLR, 2022.

\end{thebibliography}
\bibliographystyle{tmlr}

\clearpage
\appendix

\section{TSF datasets}
\label{app:tsf_datasets}
Blow we provide a detailed description of the datasets used in the paper. A summery of the datasets can be found in Tab.~\ref{tab:datasets}.

\paragraph{Electricity Transformer Temperature (ETT)~\citet{zhou2021informer}:} The ETT contains electricity power load (six features) and oil temperature collected over a period of two years from two countries in China. The dataset is versatile and exhibits short-term periodical patterns, long-term periodical patterns, long-term trends, and irregular patterns. The dataset is further divided to two granularity levels: {ETTh1, ETTh2} for one hour level and {ETTm1, ETTm2} for 15 minutes level.

\textbf{Weather\footnote{\url{ https://www.bgc-jena.mpg.de/wetter/}}:} The dataset contains 21 meteorological sensors for a range of 1 year in Germany.

\textbf{Electricity Consuming Load (ECL)\footnote{\url{https://archive.ics.uci.edu/dataset/321/electricityloaddiagrams20112014}}:} It contains the hourly electricity consumption (Kwh) of 321 clients.

\textbf{Traffic\footnote{\url{https://pems.dot.ca.gov/}}:} The dataset consists of hourly data spanning 24 months (2016-2018) obtained from the California Department of Transportation. This data provides information on road occupancy rates, measured by 862 sensors on freeways in the San Francisco Bay area, ranging between 0 and 1.

\begin{table}[ht]
    \caption{We detail several statistics regarding the datasets considered in this work.}
    \label{tab:datasets}
    \centering
    \vskip 0.1in
    \begin{tabular}{|c|c|c|c|}
    \hline
    Dataset & Number of features & Number of train samples & Granularity \\
    \hline
    ETTm1, ETTm2 & 7 & 34369 & 15 minutes \\
    \hline
    ETTh1, ETTh2 & 7 & 34369 & 1 hour \\
    \hline
    Weather& 21 & 36696  & 1 hour \\
    \hline
    ECL &321 & 18221   & 1 hour \\
    \hline
    Traffic& 862 & 12089 & 1 hour \\
    \hline
    \end{tabular}
\end{table}

\section{Additional results}
\label{app:add_results}
\subsection{ETT* datasets analysis}
\begin{figure}[t]
    \centering
    \includegraphics[width=1\linewidth]{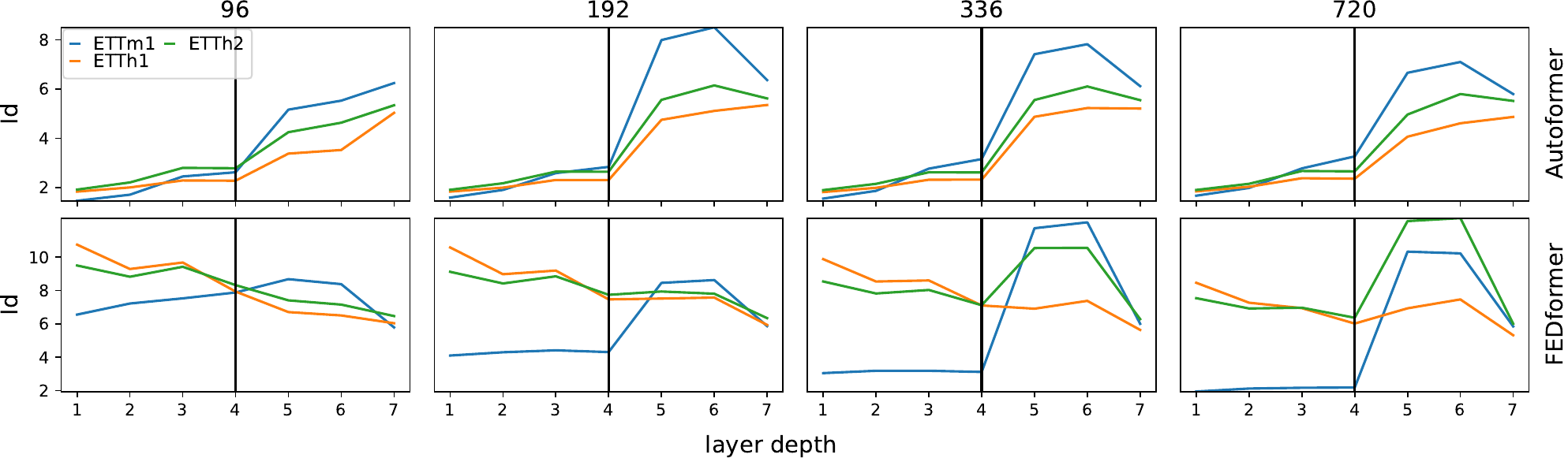}
     \caption{\textbf{ID profiles across layers of Autoformer and FEDformer on ETT datasets for multiple forecasting horizons.} Each panel includes separate ID profiles per dataset, for several horizons (left to right) and architectures (top to bottom).}
    \label{fig:id_ett}
\end{figure}

\begin{figure}[t!]
    \centering
    \includegraphics[width=1\linewidth]{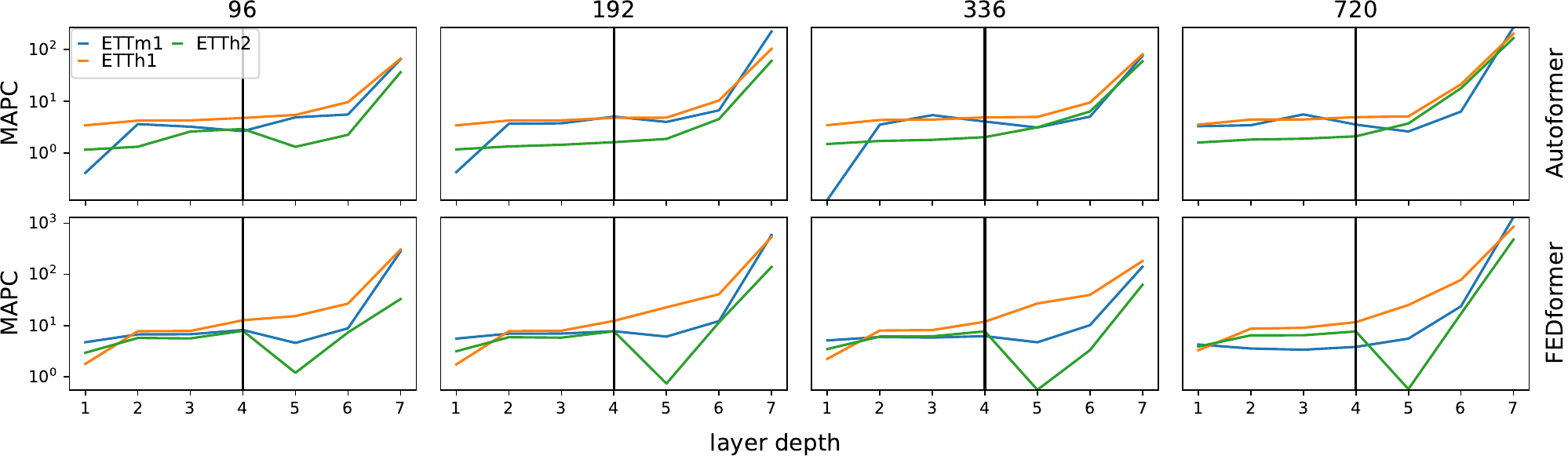}
     \caption{\textbf{MAPC profiles across layers of Autoformer and FEDformer on ETT datasets for multiple forecasting horizons.} Each panel includes separate MAPC profiles per dataset, for several horizons (left to right) and architectures (top to bottom).}
    \label{fig:mapc_ett}
\end{figure}

To complement the results in the main article we add a comparison of the ID and MAPC of three different ETT datasets: ETTm1, ETTh1 and ETTh2. We notice that ETTm1, ETTh1 and ETTh2 show a hunchback trend for Autoformer while for FEDformer the hunchback trend appears for larger forecast horizons as shown in Fig.~\ref{fig:id_ett}. Our MAPC results in Fig.~\ref{fig:mapc_ett} show that MAPC are relatively fixed and start to rise at the decoder.

\begin{figure}[b!]
    \centering
    \includegraphics[width=1\linewidth]{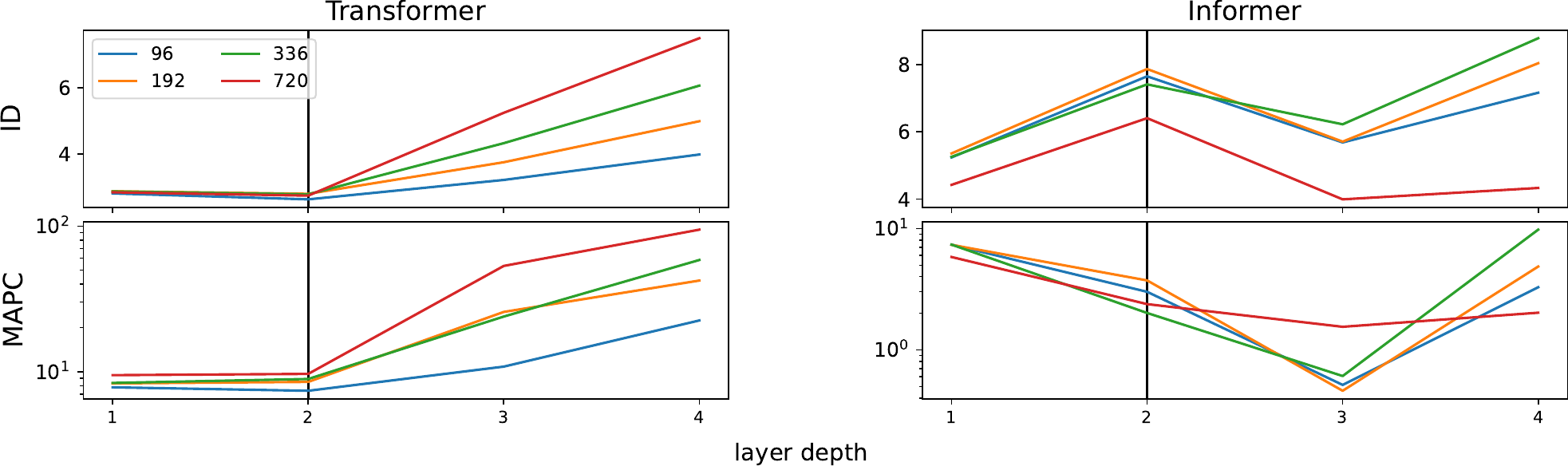}
    \caption{\textbf{Intrinsic dimension and mean absolute principal curvature along the layers of Transformer and Informer on traffic dataset for multiple forecasting horizons.}  Top) intrinsic dimension. Bottom) mean absolute principal curvature. For each model, both ID and MAPC share a similar profile across different forecasting horizons.}
    \label{fig:id_mapc_trans_inf}
\end{figure}

\section{TSF models}
\label{app:tsf_models}
Here we will provide a supplementary analysis of additional TSF models: vanilla Transformer~\citep{vaswani2017attention}, Informer~\citep{zhou2021informer} and PatchTST~\citep{nie2023time}.

\begin{figure}[t!]
    \centering
    \includegraphics[width=1\linewidth]{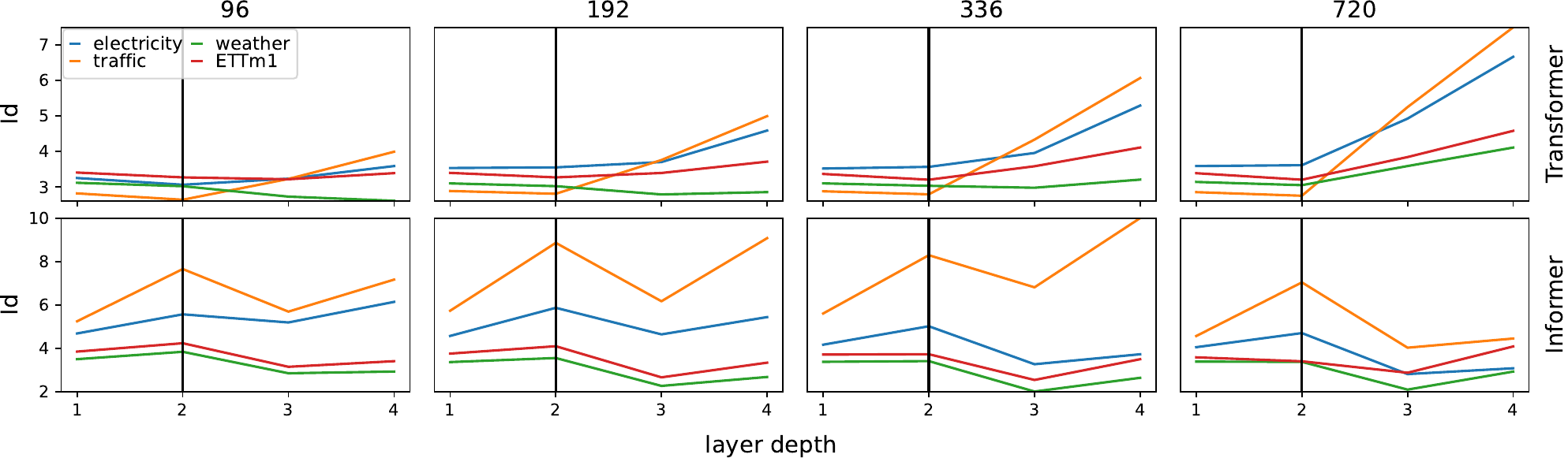}
    \caption{\textbf{ID profiles across layers of Transformer and Informer on electricity, traffic, weather and ETTm1 datasets for multiple forecasting horizons.} Each panel includes separate MAPC profiles per dataset, for several horizons (left to right) and architectures (top to bottom).}
    \label{fig:id_trans_inf_mult_datasets}
\end{figure}

\begin{figure}[b!]
    \centering
    \includegraphics[width=1\linewidth]{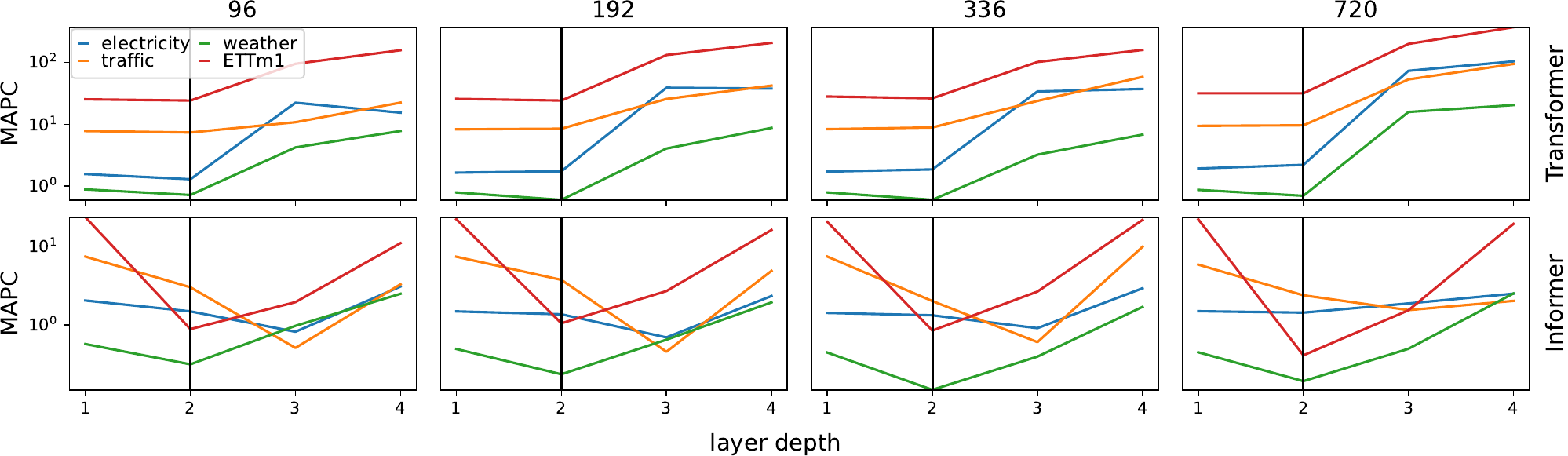}
    \caption{\textbf{MAPC profiles across layers of Transformer and Informer on electricity, traffic, weather and ETTm1 datasets for multiple forecasting horizons.} Each panel includes separate MAPC profiles per dataset, for several horizons (left to right) and architectures (top to bottom).}
    \label{fig:mapc_trans_inf_mult_datasets}
\end{figure}

\textbf{Transformer and Infromer:} These models have a similar architecture to the Autoformer and FEDformer as shown in Fig.~\ref{fig:arch}. The models are composed of two encoder layers and one decoder layer, however, in contrast to Autoformer and FEDformer, Transformer and Infromer do not contain series decomposition layers. The analysis of Transformer and Infromer inspects the output of the encoder layers, decoder layer and the last linear layer. In Fig.~\ref{fig:id_mapc_trans_inf} we observe trends similar to ones shown for the Autoformer and FEDformer. When comparing different datasets, we see similar trends, a monotonic increase in ID for the Transformer and a saw-like behaviour for Informer (see Fig.\ref{fig:id_trans_inf_mult_datasets}). The MAPC across datasets for the Transformer exhibits a monotonic increase trend while the Informer has a ``v''-shape (see Fig.\ref{fig:mapc_trans_inf_mult_datasets})

\begin{figure}[t]
    \centering
    \includegraphics[width=1\linewidth]{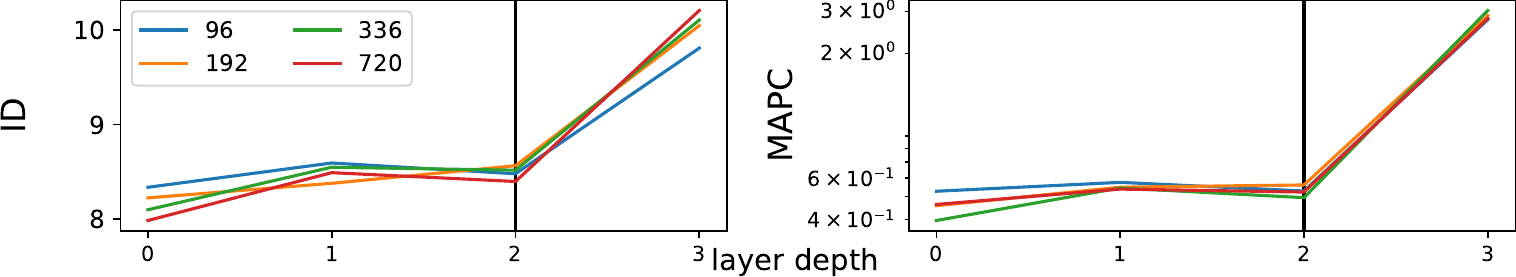}
    \caption{\textbf{ID and MAPC along the layers of PatchTST on weather dataset for multiple forecasting horizons.}  Left) intrinsic dimension. Right) mean absolute principal curvature. ID and MAPC share a similar profile across different forecasting horizons.}
    \label{fig:id_mapc_tst}
\end{figure}

\begin{figure}[b!]
    \centering
    \includegraphics[width=1\linewidth]{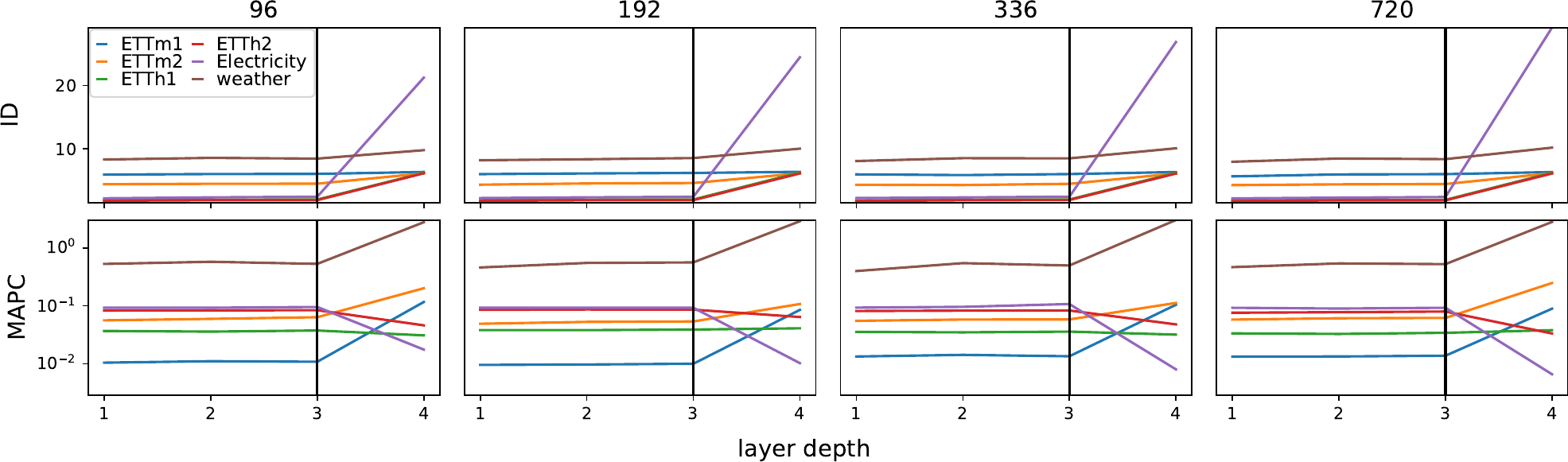}
    \caption{\textbf{ID and MAPC profiles across layers of PatchTST on ETT*, electricity and weather datasets for multiple forecasting horizons.} Each panel includes separate ID (top) and MAPC (bottom) profiles per dataset, for several horizons (left to right).}
    \label{fig:id_tst_mult_datasets}
\end{figure}

\textbf{PatchTST:} PatchTST is a transformer-based model composed of vanilla Transformer encoders and a linear layer as a decoder. PatchTST differs from all other architecture mentioned in the paper by dividing a sequence into patches and using them as inputs to the network. Each patch is then embedded, and from that point on, all sequential information becomes inaccessible, i.e., we cannot manually extract temporal information for each time stamp in the patch. Our analysis focuses on time series manifolds, where each element represents a single point in time, while the latent representation of the PatchTST model lie on product manifolds where each element is a sequence of points in time. Similar to the analysis performed on the Transformer and Informer, we inspect the output of the encoder layers and the decoder layer which is the last linear layer. We notice that the ID remains relatively stable during the encoding phase and increases in the last layer, a behavior consistent across datasets and horizons. A similar trend is observed for the MAPC during the encoding phase where on some datasets the MAPC increases on the last layer and on other it decreases. See Figs.~\ref{fig:id_mapc_tst} and \ref{fig:id_tst_mult_datasets}.

\section{Intrinsic dimension and mean absolute principal curvature}
\label{app:id_and_mapc}
\subsection{Intrinsic dimension}
To estimate the ID of data representations in TSF neural networks, we use the TwoNN~\citep{facco2017estimating} global id estimator. The ID-estimator utilizes the distances only to the first two nearest neighbors of each point. This minimal selection helps reduce the impact of inconsistencies in the dataset during the estimation process.
\paragraph{Method} Let $X=\{x_{1},x_{2}, \cdots , x_{N}\}$ a set of points uniformly sampled on a manifold with intrinsic dimension $d$. For each point $x_{i}$, we find the two shortest distances $r_{1}, r_{2}$ from elements in $X \setminus \{ x_{i} \}$ and compute the ratio $\mu_{i}=\frac{r_{2}}{r_{1}}$. It can be shown that $\mu_{i}, 1 \leq i \leq N$ follow a Pareto distribution with parameter $d + 1$ on $[1,\infty)$, that is $f\left(\mu_i \mid d\right)=d \mu_i^{-(d+1)}$. While $d$ can be estimated by maximizing the likelihood:
\begin{equation}
    \label{eqn:twonn_like}
    P(\mu_{1}, \mu_{2}, \cdots \mu_{N} \mid d)=d^N \prod_{i=1}^N \mu_i^{-(d+1)}
\end{equation}

we follow the method proposed by~\citet{facco2017estimating} based on the cumulative distribution $F(\mu)=1-\mu^{-d}$. The idea is to estimate $d$ by a linear regression on the empirical estimate of $F(\mu)$. This is done by sorting the values of $\mu$ in ascending order and defining $F^{emp}\left(\mu_{i}\right) \doteq \frac{i}{N}$. A straight line is then fitted on the points of the plane $\left\{\left(\log \mu_i,-\log \left(1-F_i^{e m p}\right)\right)\right\}_{i=1}^N$. The slope of the line is the estimated ID.

\newpage

\subsection{Comparison of intrinsic dimension estimators}
\begin{figure}[t]
    \centering
    \includegraphics[width=1\linewidth]{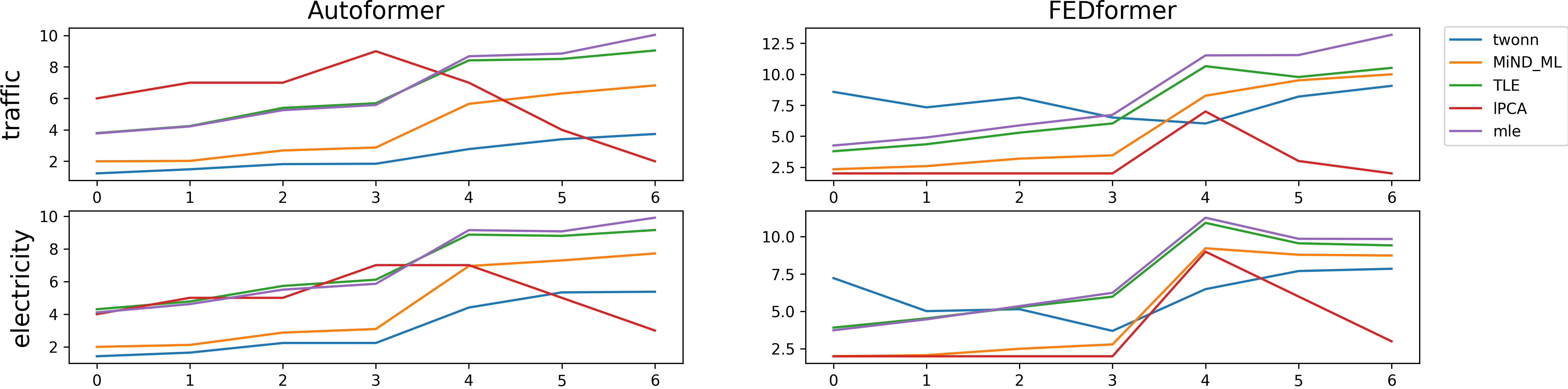}
    \caption{\textbf{Comparison of several ID estimation tools.} Most estimators agree with the trend observed by the TwoNN ID estimator.}
    \label{fig:id_comp}
\end{figure}

ID estimators operate based on a variety of principles. They are developed using specific features such as the number of data points within a fixed-radius sphere, linear separability, or the expected normalized distance to the nearest neighbor. Consequently, different ID estimation methods yield varying ID values. In Fig.~\ref{fig:id_comp} we proivde a comparison of several ID estimators from \code{Scikit-dimension}~\citep{bac2021scikit}. The results indicate that the ID profile obtained by the TwoNN estimator used in this work is simillar to other ID estimation tools. We observe a different trend from the LPCA estimator which we attribute to the linear nature of the algorithm. LPCA is grounded in the observation that for data situated within a linear subspace, the dimensionality corresponds to the count of non-zero eigenvalues of the covariance matrix. We note that the inconsistency appears in the layers where the curvature increases and thus the manifold deviates from a flat (linear) surface.

\subsection{Data density}
In comparison to the intrinsic dimension, which is a characteristic of the entire manifold, curvature information is local. Moreover, curvatures are calculated using second-order derivatives of the manifold. Consequently, our study assumes that the data is dense enough to compute curvatures. However, the latent representations of data, are both high-dimensional and sparse, which presents significant difficulties in calculating local differentiable values on such as curvature.

The typical characteristics of data used in machine learning require a large number of nearby points to create a stable neighborhood. One commonly used tool for this is k-Nearest-Neighbours (KNN). However, KNN can sometimes generate non-local and sparse neighborhoods, where the "neighbors" are effectively far apart in a Euclidean sense. Another approach is to use domain-specific augmentations, such as window cropping, window warping or slicing. However, this approach only explores a specific aspect of the data manifold and may overlook other important parts. A more effective approach, regardless of the domain, is to compute the Singular Value Decomposition (SVD) for each time series. This generates a close neighborhood by filtering out small amounts of noise in the data. This approach is well-motivated from a differential geometry standpoint, as it approximates the manifold at a point and samples the neighborhood.

\paragraph{Neighborhood generation.} To improve the local density of time sereis samples, we use a procedure similar to~\citet{yu2018curvature} to generate artificial new samples by reducing the ``noise'' levels of the original data. Specifically, given a $d$ dimensional time series $x_{1:T}\in \mathbb{R}^{T \times d}$, let $x_{1:T}=U \Sigma V^T$ be its SVD, where $U\in \mathbb{R}^{T \times T}$, $V\in \mathbb{R}^{d \times d}$ and $\Sigma \in \mathbb{R}^{T \times d}$ a rectangular diagonal matrix with singular values $\left\{\sigma_{1}, \sigma_{2}, \cdots, \sigma_{d} \right\}$ on the diagonal in descending order such that $r$ is the rank of $x_{1:T}$. Let $m$ be the smallest index such that the explained varince $\frac{\sigma_m^2}{\sum_j \sigma_j^2}$ of the $m$'th mode is less than or equal to $1e^{-3}$. We define $\Sigma'$=$\left\{\sigma_{1}, \sigma_{2}, \cdots, u_{1}\sigma_{m}, u_{2}\sigma_{m+1}, \cdots u_{d-m+1}\sigma_{d} \right\} $ such that $u_{i} \stackrel{\text { i.i.d. }}{\sim} \mathbb{U}(0,1)$ and construct $x_{1:T}'=U \Sigma ' V^T$. this process is repeated $64$ times for each time series, generating $64$ new time series.  
\label{app:id_est}

\subsection{Curvature estimation}
There are several methods available for estimating curvature quantities of data representations, as discussed in papers such as~\citet{brahma2015deep,shao2018riemannian}. For our purposes, we have chosen to use the algorithm described in~\citet{li2018curvature}, which is called Curvature Aware Manifold Learning (CAML). We opted for this algorithm because it is supported by theoretical foundations and is relatively efficient. In order to use CAML, we need to provide the neighborhood information of a sample and an estimate of the unknown ID. The ID is estimated using the TwoNN algorithm, as described in~\ref{app:id_est}, similarly to~\citet{ansuini2019intrinsic,kaufman2023data}.

In order to estimate the curvature of data $Y=\{y_1, y_2, \cdots, y_N \}\subset\mathbb{R}^{D}$, we make the assumption that the data lies on a $d$-dimensional manifold $\mathcal{M}$ embedded in $\mathbb{R}^{D}$, where $d$ is much smaller than $D$. Consequently, $\mathcal{M}$ can be considered as a sub-manifold of $\mathbb{R}^{D}$. The main concept behind CAML is to compute a local approximation of the embedding map using second-order information.
\begin{equation}
    \label{EQN:embmap}
    f: \mathbb{R}^d \rightarrow \mathbb{R}^D \quad, y_i = f(x_i) + \epsilon_{i} \ , \quad i=1,\dots,N \ ,
\end{equation}
where $X=\{x_1, x_2, \cdots, x_N\}\subset\mathbb{R}^{d}$ are low-dimensional representations of $Y$, and $\{\epsilon_1, \epsilon_2, \cdots \epsilon_N\}$ are the noises. In the context of this paper, the embedding map $f$ is the transformation that maps the low-dimensional dynamics to the sampled features for each time stamp $t$ that might hold redundant information.

In order to estimate curvature information at a point $y_i \in Y$, we follow the procedure described above to define its neighborhood. This results in a set of nearby points $\{ y_{i_1}, \dots, y_{i_K} \}$, where $K$ represents the number of neighbors. Using this set along with the point $y_i$, we utilize SVD to construct a local natural orthonormal coordinate frame $\left\{\frac{\partial}{\partial x^1}, \cdots, \frac{\partial}{\partial x^d}, \frac{\partial}{\partial y^1}, \cdots, \frac{\partial}{\partial y^{D-d}}\right\}$. This coordinate frame consists of a basis for the tangent space (first $d$ elements) and a basis for the normal space.
To be precise, we denote the projection of $y_i$ and $y_{i_j}$ for $j=1, \dots, K$ onto the tangent space spanned by $\partial/\partial x^1, \dots, \partial/\partial x^d$ as $x_i$ and $u_{i_j}$ respectively. It is important to note that the neighborhood of $y_i$ must have a rank of $r>d$. If the rank is less than $d$, then SVD cannot accurately encode the normal component at $x_i$, leading to poor approximations of $f$ at $x_i$. Therefore, we verify that $\{y_{i_1}, \dots, y_{i_K} \}$ has a rank of $d+1$ or higher.

The map $f$ can be expressed in the alternative coordinate frame as $f(x^1,\dots,x^d) = [x^1, \dots, x^d, f^1, \dots, f^{D-d}]$. The second-order Taylor expansion of $f^\alpha$ at $u_{i_j}$ with respect to $x_i$, with an error of $\mathcal{O}(|u_{i_j}|_2^2)$, is represented by
\begin{equation}
    \label{EQN:maptaylor}
    f^\alpha(u_{i_j}) \approx f^\alpha(x_i) + \Delta _{x_i}^T \nabla f^\alpha +\frac{1}{2} \Delta _{x_i}^T H^\alpha \Delta _{x_i} \ ,
\end{equation}
where $\alpha=1,\dots,D-d$, $\Delta _{x_i}=(u_{i_j}-x_i)$ and $u_{i_j}$ is an elemnt in the neighborhood of $x_i$. The gradient of $f^\alpha$ is denoted by $\nabla f^\alpha$, and $H^\alpha = \left( \frac{\partial^2 f^\alpha}{\partial x^i \partial x^j} \right)$ is its Hessian. We have a neighborhood $\{ y_{i_1}, \dots, y_{i_K}\}$ of $y_i$, and their corresponding tangent representations $\{ u_{i_j} \}$. Using equation~\ref{EQN:maptaylor}, we can form a system of linear equations, as explained in~\ref{app:taylor_linsolve}. The principal curvatures are the eigenvalues of $H^\alpha$, so estimating curvature information involves solving a linear regression problem followed by an eigendecomposition. Each Hessian has $d$ eigenvalues, so each sample will have $\left(D-d \right) \times d$ principal curvatures. Additionally, one can compute the Riemannian curvature tensor using the principal curvatures, but this requires high computational resources due to its large number of elements. Moreover, as the Riemannian curvature tensor is fully determined by the principal curvatures, we focus our analysis on the eigenvalues of the Hessian. To evaluate the curvature of manifolds, we estimate the mean absolute principal curvature (MAPC) by taking the mean of the absolute values of the eigenvalues of the estimated Hessian matrices.

\subsection{Estimating the Hessian Matrix}
\label{app:taylor_linsolve}
In order to estimate the Hessian of the embedding mapping $f^\alpha$ where $\alpha=1,\dots,D-d$, we build a set of linear equations that solves Eq.~\ref{EQN:maptaylor}. We approximate $f^\alpha$ by solving the system $f^\alpha = \Psi X_{i}$, where $X_{i}$ holds the unknown elements of the gradient $\nabla f^\alpha$ and the hessian $H^\alpha$. We define $f^\alpha=\left[f^\alpha\left(u_{i_1}\right), \cdots, f^\alpha\left(u_{i_K}\right)\right]^T$, where $u_{i_j}$ are points in the neighborhood of $x_i$, in the local natural orthogonal coordinates. The local natural orthogonal coordinates are a set of coordinates that are defined at a specific point $p$ of the manifold. They are constructed by finding a basis for the tangent space and normal space at a point $p$ by applying  Principal Component Analysis, such that the first $d$ coordinates (associated with the most significant modes, i.e., largest singular values) represent the tangent space, and the rest represent the normal space. We define $\Psi=\left[\Psi_{i_1}, \cdots, \Psi_{i_K}\right]$, where $\Psi_{i_j}$ is given via
\begin{equation*}
    \Psi_{i_j}=\left[u_{i_j}^1, \cdots, u_{i_j}^d,\left(u_{i_j}^1\right)^2, \cdots,\left(u_{i_j}^d\right)^2,\left(u_{i_j}^1 \times u_{i_j}^2\right), \cdots,\left(u_{i_j}^{d-1} \times u_{i_j}^d\right)\right] \ .
\end{equation*}

The set of linear equations $f^\alpha=\Psi X_{i}$  is solved by using the least square estimation resulting in $X_{i}=\Psi^{\dagger}f^\alpha$, where $X_{i}=\left[{\nabla f^\alpha}^1,\cdots, {\nabla f^\alpha}^d, {H^\alpha}^{1,1}, \cdots, {H^\alpha}^{d,d}, {H^\alpha}^{1,2}, \cdots, {H^\alpha}^{d-1,d}\right]$. In practice, we estimate only the upper triangular part of ${H^\alpha}$ since it is a symmetric matrix. The gradient values $\nabla f^\alpha$ are ignored since they are not required for the CAML algorithm. We refer the reader for a more comprehensive and detailed analysis in~\citep{li2018curvature}.

\end{document}